\documentclass{article}

 \usepackage[preprint]{neurips_2025}

\usepackage[utf8]{inputenc} % allow utf-8 input
\usepackage[T1]{fontenc}    % use 8-bit T1 fonts
\usepackage{hyperref}       % hyperlinks
\usepackage{url}            % simple URL typesetting
\usepackage{booktabs}       % professional-quality tables
\usepackage{amsfonts}       % blackboard math symbols
\usepackage{nicefrac}       % compact symbols for 1/2, etc.
\usepackage{microtype}      % microtypography
\usepackage{xcolor}         % colors

\usepackage{graphicx}%
\usepackage{multirow}%
\usepackage{amsmath,amssymb,amsfonts}%
\usepackage{amsthm}%
\usepackage{mathrsfs}%
\usepackage[title]{appendix}%
\usepackage{xcolor}%
\usepackage{textcomp}%
\usepackage{manyfoot}%
\usepackage{booktabs}%
\usepackage{algorithm}%
\usepackage{algorithmicx}%
\usepackage{listings}%
\usepackage{graphicx}
\usepackage{amssymb}
\usepackage{arydshln}
\usepackage{booktabs}
\usepackage{multirow}
\usepackage{caption}
\usepackage{subcaption}
\usepackage{makecell}
\usepackage{csquotes}
\usepackage{epigraph}
\usepackage{amsmath}
\usepackage{xcolor}  
\usepackage{cleveref}
\usepackage{wrapfig}
\usepackage{pifont}
\usepackage{listings}
\usepackage{appendix}
\usepackage{algorithm}
\usepackage{multicol}
\usepackage{mathtools}
\usepackage{placeins}
\usepackage{algorithm}    % for algorithm environment
\usepackage[noend]{algpseudocode}

\algblock[noend]{ParFor}{EndParFor}
\algnewcommand\algorithmicparfor{\textbf{parfor}}
\algnewcommand\algorithmicpardo{\textbf{do}}
\algnewcommand\algorithmicendparfor{\textbf{end\ parfor}}
\algrenewcommand\alglinenumber[1]{}
\algrenewtext{ParFor}[1]{\algorithmicparfor\ #1\ \algorithmicpardo}
\algrenewtext{EndParFor}{\algorithmicendparfor}
\algtext*{EndParFor}

\algnewcommand\algorithmicinput{\textbf{Define}}
\algnewcommand\Def{\item[\algorithmicinput]}

\newcommand\net{\text{MatMul-free LM}}
\newcommand\name{\text{MLGRU}}

\title{Scalable MatMul-free Language Modeling}

\author{%
  Rui\mbox{-}Jie Zhu$^1$, Yu Zhang$^2$, Steven Abreu$^3$\thanks{Work done during an internship at Intel Labs.}\ , Ethan Sifferman$^1$, Tyler Sheaves$^1$, \\
  \textbf{Yiqiao Wang}$^4$, \textbf{Dustin Richmond}$^1$, \textbf{Sumit Bam Shrestha}$^5$, \textbf{Peng Zhou}$^{14}$, \textbf{Jason K.\ Eshraghian}$^1$\thanks{Correspondence to \texttt{jsn@ucsc.edu}.}\\[0.1cm]
  $^1$ University of California, Santa Cruz, $^2$ Soochow University, $^3$ University of Groningen,\\
  $^4$ LuxiTech, $^5$ Intel Labs
}

\begin{document}

\maketitle

\begin{abstract}
Large Language Models (LLMs) have fundamentally altered how we approach scaling in machine learning. However, these models pose substantial computational and memory challenges, primarily due to the reliance on matrix multiplication (MatMul) within their attention and feed-forward (FFN) layers. We demonstrate that MatMul operations can be eliminated from LLMs while maintaining strong performance, even at billion-parameter scales. Our MatMul-free models, tested on models up to 2.7B parameters, are comparable to state-of-the-art pre-trained Transformers, and the performance gap narrows as model size increases.

Our approach yields significant memory savings: a GPU-efficient implementation reduces memory consumption by up to 61\% during training and over 10$\times$ during inference. When adapted for a multi-chip neuromorphic system, the model leverages asynchronous processing to achieve 4$\times$ higher throughput with 10$\times$ less energy than edge GPUs. %and 77$\times$ less energy than server-class GPUs, demonstrating superior scaling. 
These findings demonstrate a path toward dramatically simplified yet effective LLMs, advancing them toward brain-like efficiency and heralding a new generation of lightweight, high-performance language models.
Our code implementation is available at \href{https://github.com/ridgerchu/matmulfreellm}{https://github.
com/ridgerchu/matmulfreellm}.
\end{abstract}

\section{Main}

% Matrix Multiplication (MatMul) is a fundamental operation in neural networks, playing a crucial role in various components.
Matrix Multiplication (MatMul) is the dominant operation in most neural networks, where dense layers involve vector-matrix multiplication (VMM), convolutions can be implemented as block-sparse VMMs with shared weights, and self-attention relies on matrix-matrix multiplication (MMM).
% In fully-connected dense layers, MatMul is used for functioning as the weight, mimicking the synapse operations of the human brain. In self-attention layers, MatMul is employed to simulate the human attention mechanism. 
% Even convolution layers, in their practical implementation, rely on MatMul for acceleration. 
The prevalence of MatMul is primarily due to Graphics Processing Units (GPUs) being optimized for MatMul operations. By leveraging Compute Unified Device Architecture (CUDA) and highly optimized linear algebra libraries such as cuBLAS, the MatMul operation can be efficiently parallelized and accelerated. This optimization was a key factor in the victory of AlexNet in the ILSVRC2012 competition and a historic marker for the rise of deep learning~\cite{krizhevsky2012imagenet}. AlexNet notably utilized GPUs to boost training speed beyond CPU capabilities, and as such, deep learning won the `hardware lottery'~\cite{hooker2021hardware}. It also helped that both training and inference rely on MatMul.

% Secondly, MatMul's compatibility with backpropagation makes it a practical choice for model training~\cite{rumelhart1986learning}.

Despite its prevalence in deep learning, MatMul operations account for the dominant portion of computational expense, often consuming the majority of the execution time and memory access during both training and inference phases.  
Several works have replaced MatMul with simpler operations through two main strategies. The first strategy involves substituting MatMul with elementary operations, e.g., AdderNet replaces multiplication with signed addition in convolutional neural networks (CNNs)~\cite{chen2020addernet}. Given the focus on convolutions, AdderNet is intended for use in computer vision over language modeling. ShiftAddLLM enables re-parameterization of dense layers, though attention layers still rely on dynamic MatMul operations~\cite{you2024shiftaddllm}, where I/O limits will quickly overtake arithmetic savings as the sequence length grows.

The second approach employs binary or ternary quantization, simplifying MatMul to operations where values are either flipped or zeroed out before accumulation. 
Quantization can be applied to either activations or weights: spiking neural networks (SNNs) use binarized activations~\cite{maass1997networks, eshraghian2023training, zhu2023spikegpt}, while binary and ternary neural networks (BNNs and TNNs) use quantized weights~\cite{courbariaux2016binarized}. Both methods can also be combined~\cite{venkatesh2024squat, eshraghian2022memristor}.

Recent advances in language modeling, like BitNet and Falcon-Edge~\cite{wang2023bitnet, ma20241bitllm, tiionebitllms}, demonstrate quantization's scalability, replacing all dense layer weights with binary/ternary values to support up to 3B parameters.
Despite replacing VMMs with accumulations in all dense layers, they retain the self-attention mechanism which relies on an expensive MMM. Dynamically computed matrices $Q$ (query) and $K$ (key) are multiplied to form the attention map. Since both $Q$ and $K$ matrices are dynamically computed from pre-activation values, achieving optimal hardware efficiency on GPUs requires custom optimizations, such as specialized kernels and advanced memory access patterns. Despite these efforts, such MatMul operations remain resource-intensive on GPUs, as they involve extensive data movement and synchronization which can significantly hinder computational throughput and efficiency~\cite{dao2022flashattention}. In our experiments, ternary quantization of the attention matrices in BitNet causes a significant drop in performance and failure to reach model convergence (see Fig.~\ref{fig:training_loss}). This raises the question: is it possible to completely eliminate MatMul from LLMs?

In this work, we develop the first scalable MatMul-free language model (Matmul-free LM) by using additive operations in dense layers and element-wise Hadamard products for self-attention-like functions. Specifically, ternary weights eliminate MatMul in dense layers, similar to BNNs. To remove MatMul from self-attention, we optimize the Gated Recurrent Unit (GRU)~\cite{cho2014learning} to rely solely on element-wise products and show that this model competes with state-of-the-art Transformers while eliminating all MatMul operations. 

To fully exploit the efficiency potential of our architecture, we developed microcode-level optimization to process this model on a neuromorphic cluster using Intel's Loihi 2 platform. The architecture of our MatMul-free LM naturally aligns with neuromorphic computing paradigms, allowing us to achieve remarkable efficiency gains. By mapping our model to Loihi 2's mesh of asynchronous neurocores, we surpass human-readable throughput by approximately 8$\times$ at 4.2~W power consumption, representing a significant advancement over conventional hardware. 
%During prefill operations, our implementation processes tokens at up to 6,632 tokens/second while consuming 2.8 mJ/token, outperforming comparable Transformer models on embedded GPUs by at least 4$\times$ in throughput and 3$\times$ in energy efficiency. 
For autoregressive generation, our model maintains consistent 59.4 tokens/second throughput at a highly efficient 70.8 mJ/token, far outperforming comparable Transformer models on embedded GPUs by at least 4$\times$ in throughput and 10$\times$ in energy efficiency, demonstrating how neuromorphic hardware can effectively harness the MatMul-free properties of our architecture. This implementation moves LLMs closer to brain-like efficiency and points towards a new generation of lightweight, high-performance language models.

\section{Building the MatMul-free Language Model}
\begin{figure}
    \centering
    \includegraphics[width=\textwidth]{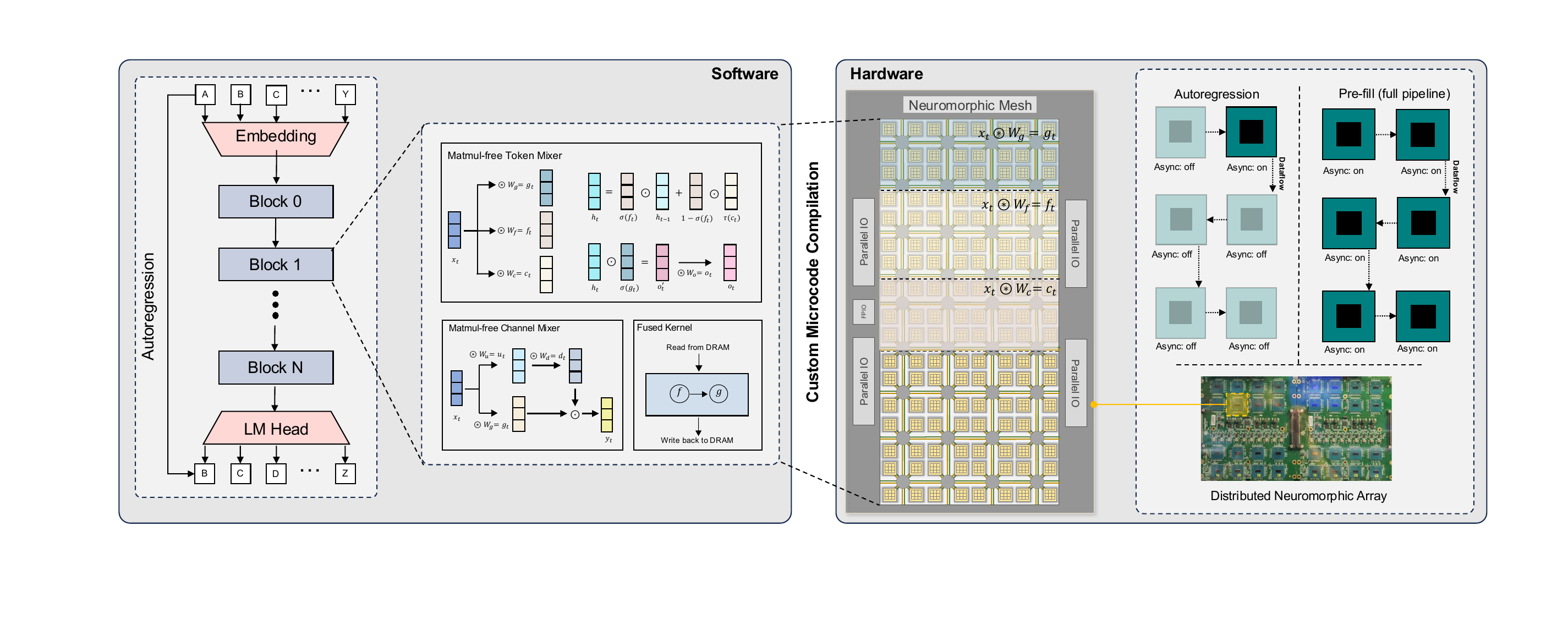}
    \caption{Overview of the \net{}. Left: general architecture of proposed model. Middle-right: Algorithm mapping of a single block across neurocores on a single Loihi2 chip. Top-Right: Multi-chip Dataflow for autoregression and pre-fill. 
    During autoregression, only one chip consumes non-negligible dynamic power dissipation due to the clock-free system. Bottom-right: The \net{} is deployed on the Hala Point system which consists of 1,152 Loihi~2 chips.}
    \label{fig:main}
\end{figure}

A closer examination of the Transformer architecture reveals two fundamental components that heavily rely on MatMul operations. The first is the dense layer, which not only transforms input hidden states to create Query, Key, and Value matrices for attention computation but also serves as the core structure in Feed-Forward Network (FFN) layers. Indeed, dense layers account for nearly all parameters in Transformer models. The second MatMul-dependent component is the attention mechanism itself, which performs MatMul operations directly on activations to compute attention scores.

The challenge, therefore, is to eliminate MatMul operations from both these structures while maintaining comparable performance. Following the Metaformer~\cite{yu2022metaformer} framework, we can decompose the Transformer architecture into two essential functions: token mixing and channel mixing. The token mixer operates on input sequences (implemented as attention in traditional Transformers), while the channel mixer processes information across embedding dimensions (implemented as FFN layers in Transformers).

Our solution addresses both components through a comprehensive architectural transformation. We replace the attention mechanism with an element-wise recurrent neural network (RNN) that provides similar token mixing capabilities but relies solely on element-wise operations. Simultaneously, we transform all dense layers to use ternary weights, effectively eliminating MatMul operations throughout the entire architecture. This approach maintains the core functional properties of Transformers while dramatically reducing computational complexity.
This architectural transformation achieves a complete elimination of MatMul operations through carefully designed structural modifications, while preserving the essential mixing capabilities that make Transformers effective. Specific implementation details are provided in \ref{sec:methods}. The following sections detail how each component is optimized and demonstrate the effectiveness of this MatMul-free approach.

% \section{Scaling Properties of MatMul-free Language Models} 
\section{Scaling Analysis}
\begin{figure}
    \centering
    \includegraphics[width=0.9\textwidth]{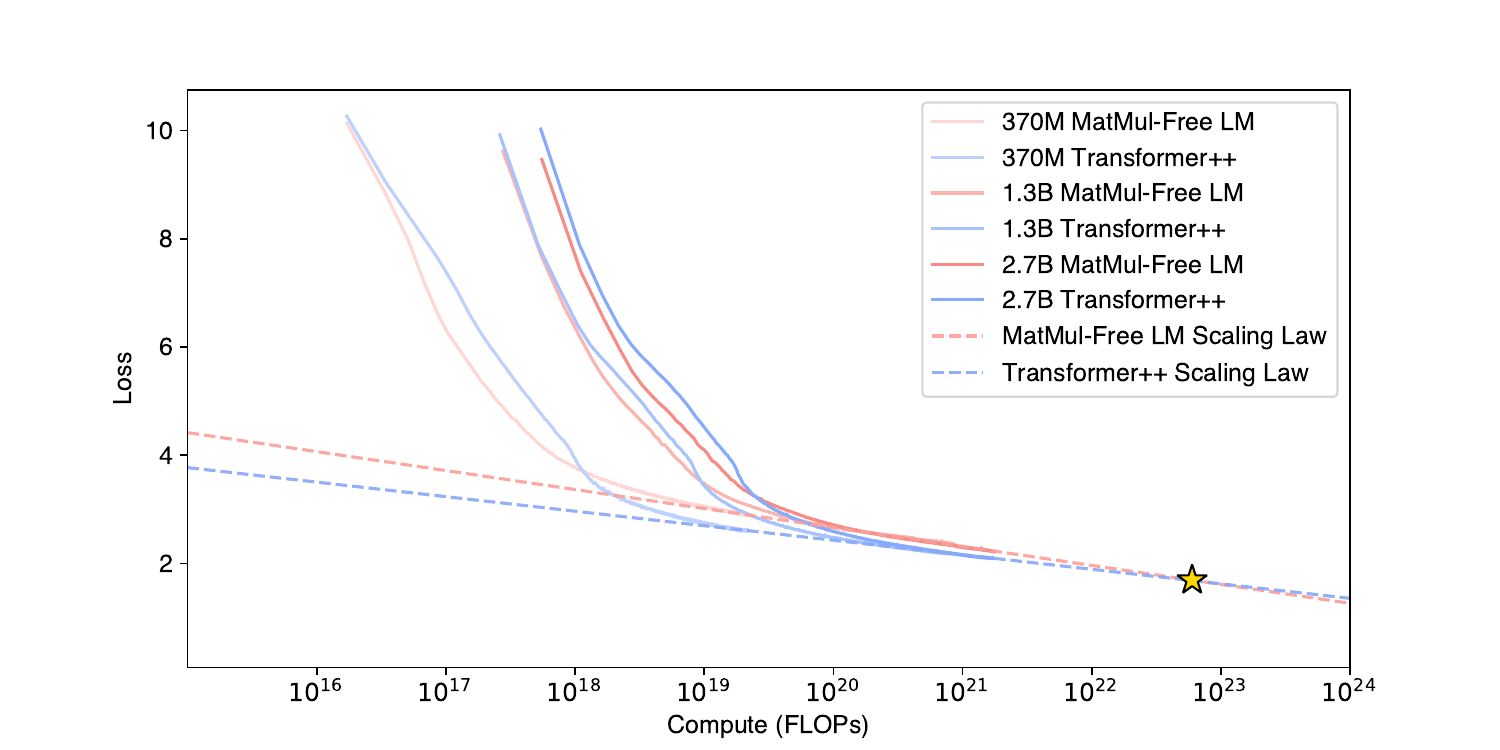}
    \caption{Scaling law comparison between \net{} and Transformer++ models, depicted through their loss curves. The red lines represent the loss trajectories of the \net{}, while the blue lines indicate the losses of the Transformer++ models. The star marks the intersection point of the scaling law projection for both model types. \net{} uses ternary parameters and BF16 activations, whereas Transformer++ uses BF16 parameters and activations.}
    \label{fig:scaling law}
\end{figure}

Neural scaling laws posit that model error decreases as a power function of training set size and model size. Such projections become important as training becomes increasingly expensive with larger models. A widely adopted best practice in LLM training is to first test scalability with smaller models, where scaling laws begin to take effect~\cite{scalinglawopenai, chichinllascalinglaw, yang2023baichuan}. The GPT-4 technical report revealed that a prediction model just $1/10,000$ the size of the final model can still accurately forecast the full-sized model performance~\cite{gpt4}.

We evaluate how the scaling law fits to the 370M, 1.3B and 2.7B parameter models in both Transformer++ and \net{}, shown in Fig.~\ref{fig:scaling law}.
For a conservative comparison, each operation is treated identically between \net{} and Transformer++.  But note that all weights and activations in Transformer++ are in BF16, while BitLinear layers in \net{} use ternary parameters, with BF16 activations. As such, an average operation in \net{} will be computationally cheaper than that of Transformer++.

Interestingly, the scaling projection for the \net{} exhibits a steeper descent compared to that of Transformer++. This suggests that the \net{} is more efficient in leveraging additional compute resources to improve performance. As a result, the scaling curve of the \net{} is projected to intersect with the scaling curve of Transformer++ at approximately $10^{23}$ FLOPs. This compute scale is roughly equivalent to the training FLOPs required for Llama-3 8B (trained with 15 trillion tokens) and Llama-2 70B (trained with 2 trillion tokens), suggesting that \net{} not only outperforms in efficiency, but can also outperform in terms of loss when scaled up. 

\section{Performance on Downstream Tasks}
\label{sec:performance-downstream-tasks}
In line with benchmarking in BitNet, we evaluated the zero-shot performance of these models on a range of language tasks, including ARC-Easy~\cite{arc}, ARC-Challenge~\cite{arc}, Hellaswag~\cite{hellaswag}, Winogrande~\cite{winoGrande}, PIQA~\cite{piqa}, and OpenbookQA~\cite{openbookqa}. The results are shown in Tab.~\ref{tab:zero-shot}. All evaluations are performed using the LM evaluation harness~\cite{eval-harness}. The \net{} models achieve competitive performance compared to the Transformer++ baselines across all tasks, demonstrating its effectiveness in zero-shot learning despite the absence of MatMul operations, and the lower memory required from ternary weights. Notably, the 2.7B \net{} model outperforms its Transformer++ counterpart on ARC-Challenge and OpenbookQA, while maintaining comparable performance on the other tasks. As the model size increases, the performance gap between \net{} and Transformer++ narrows, which is consistent with the scaling law.
These results highlight that MatMul-free architectures are capable of achieving strong zero-shot performance on a diverse set of language tasks, ranging from question answering and commonsense reasoning to physical understanding.

\begin{table*}[t]
\setlength{\tabcolsep}{6pt}
\centering
\caption{Zero-shot accuracy of \net{} and Transformer++ on benchmark datasets.}
\resizebox{\textwidth}{!}{
\begin{tabular}{lccccccccc}
\toprule
\textbf{Models} & \textbf{Size} & \textbf{ARCe} & \textbf{ARCc} & \textbf{HS} & \textbf{OQ} & \textbf{PQ} & \textbf{WGe} & \textbf{Avg.} \\
\midrule
\multicolumn{9}{l}{\emph{370M parameters with 15B training tokens, Layer=24, d=1024}}\\
Transformer++ & 370M & 45.0 & 24.0 & 34.3 &  29.2 & 64.0 & 49.9 & 41.1\\
MatMul-free RWKV-4 & 370M & 44.7 & 22.8 & 31.6 & 27.8 & 63.0 &  50.3 & 40.0\\
\textbf{Ours} & 370M & 42.6 & 23.8 & 32.8 & 28.4 & 63.0 &  49.2 & 40.3\\

\midrule
\multicolumn{9}{l}{\emph{1.3B parameters with 100B training tokens, Layer=24, d=2048}}\\
Transformer++ & 1.3B & 54.1 & 27.1 & 49.3 &  32.4 & 70.3 & 54.9 & 48.0\\
MatMul-free RWKV-4 & 1.3B & 52.4 & 25.6 & 45.1 & 31.0 & 68.2 &  50.5 & 45.5\\
\textbf{Ours} & 1.3B & 54.0 & 25.9 & 44.9 & 31.4 & 68.4 &  52.4 & 46.2\\

\midrule
\multicolumn{9}{l}{\emph{2.7B parameters with 100B training tokens, Layer=32, d=2560}}\\
Transformer++ & 2.7B & 59.7 & 27.4 & 54.2 &  34.4 & 72.5 & 56.2 & 50.7\\
\textbf{Ours} & 2.7B & 58.5 & 29.7 & 52.3 & 35.4 & 71.1 &  52.1 & 49.9\\
\bottomrule

\end{tabular}
}
\label{tab:zero-shot}
\end{table*}
\begin{figure}
    \centering
    \includegraphics[width=\linewidth]{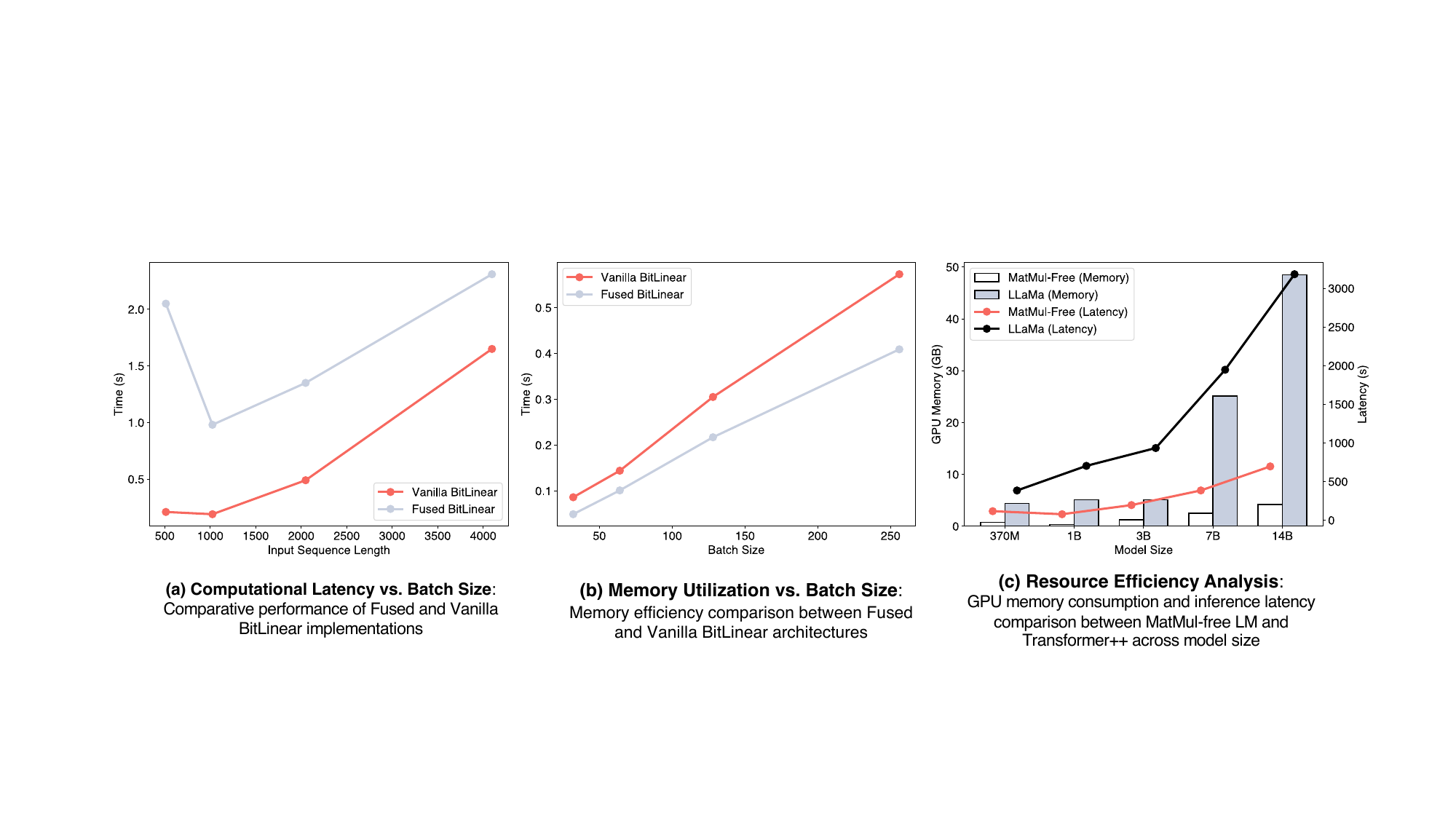}
    \caption{Performance comparison and analysis of different models and configurations. (a) and (b) show the training performance comparison between Vanilla BitLinear and Fused BitLinear in terms of time and memory consumption as a function of batch size. (c) compares the inference memory consumption and latency between MatMul-free LM and Transformer++ across various model sizes.}

    \label{fig:merge_result}
\end{figure}

% \section{Computational Efficiency and Hardware Benefits}
\section{Deployment on Neuromorphic Hardware}
\subsection{Training Efficiency Optimization on GPU}

\label{sec:exp-fused}
We evaluate our proposed Fused BitLinear and Vanilla BitLinear implementations in terms of training time and memory usage, shown in Fig.~\ref{fig:merge_result}(a-b). For each experiment, we set the input size and sequence length to 1024. All experiments are conducted using an NVIDIA A100 80GB GPU. Note that during training, the sequence length and batch dimensions are flattened, making the effective batch size the product of these dimensions.

Our experiments show that our fused operator benefits from larger batch sizes in terms of faster training speeds and reduced memory consumption. When the batch size is $2^8$, the training speed of the 1.3B parameter model improves from 1.52s to 1.21s per iteration, a 25.6\% speedup over the vanilla implementation. Additionally, memory usage decreases from 82GB to 32GB, a 61.0\% reduction. The performance of the fused implementation improves significantly with larger batch sizes, allowing more samples to be processed simultaneously and reducing the total number of iterations.

\subsection{Inference Efficiency Optimization on GPU}
Fig.~\ref{fig:merge_result}(c) presents a comparison of GPU inference memory consumption and latency between the proposed MatMul-free LM and Transformer++ for various model sizes. In the MatMul-free LM, we employ BitBLAS~\cite{bitblas} for acceleration to further improve efficiency. The evaluation is conducted with a batch size of 1 and a sequence length of 2048. The MatMul-free LM consistently demonstrates lower memory usage and latency compared to Transformer++ across all model sizes. For a single layer, the MatMul-free LM requires only 0.12 GB of GPU memory and achieves a latency of 3.79 ms, while Transformer++ consumes 0.21 GB of memory and has a latency of 13.87 ms. As the model size increases, the memory and latency advantages of the MatMul-free LM become more pronounced. It is worth noting that for model sizes larger than 2.7B, the results are simulated using randomly initialized weights. For the largest model size of 13B parameters, the MatMul-free LM uses only 4.19 GB of GPU memory and has a latency of 695.48 ms, whereas Transformer++ requires 48.50 GB of memory and exhibits a latency of 3183.10 ms. These results highlight the efficiency gains achieved by the MatMul-free LM, making it a promising approach for large-scale language modeling tasks, particularly during inference.

\subsection{Neuromorphic Computing with Intel Loihi 2}
\label{sec:loihi_results}
% We have demonstrated the feasibility and effectiveness of the first scalable MatMul-free language model. 
% Our work challenges the paradigm that MatMul operations are indispensable for building high-performing language models and paves the way for the development of more efficient and hardware-friendly architectures.
% We achieve performance on par with state-of-the-art Transformers while eliminating the need for MatMul operations, with an optimized implementation that significantly enhances both training and inference efficiency, reducing both memory usage and latency. 
% As the demand for deploying language models on various platforms grows, \net{}s present a promising direction for creating models that are both effective and resource-efficient. 
% However, one limitation of our work is that the \net{} has not been tested on extremely large-scale models (e.g., 100B+ parameters) due to computational constraints. 
% This work serves as a call to action for institutions and organizations that have the resources to build the largest language models to invest in accelerating lightweight models. By prioritizing the development and deployment of MatMul-free architectures such as this one, the future of LLMs will only become more accessible, efficient, and sustainable.
While GPUs excel at conventional deep learning workloads grounded in dense floating-point matrix multiplications, the MatMul-free LM we present is dominated by low-precision element-wise operations. The arithmetic intensity is so low that many CUDA cores remain idle during inference.
% This makes the MatMul-free LM memory-bound when using low batch sizes. 
Additionally, using ternary weights in matrix multiplications naturally induces unstructured sparsity of $\approx 35\%$ (for the 370M model) which cannot easily be accelerated on GPUs, resulting in redundant calculations and memory movement.
Our MatMul-free LM maps naturally onto large-scale neuromorphic hardware that executes element-wise, recurrent, low-precision operations close to memory. Intel’s Loihi 2, for instance, offers microcode-programmable neuron dynamics and on-chip support for sparse low-precision arithmetic, making it an ideal substrate for the MatMul-free model~\cite{davies_advancing_2021}.
% However, the computational structure of our MatMul-free LM aligns with emerging neuromorphic architectures that are designed for element-wise recurrence, fine-grained parallelism, near-memory computing and low bit precision. 
% In particular, Intel’s Loihi 2 platform \cite{davies_advancing_2021} provides flexible microcode programmability of neuron dynamics and hardware support for low-precision, sparse computations, making it an ideal match for our multiplication-free approach.
Processing is done locally within 120 fully asynchronous ``neurocores'' on each chip, with the option of connecting up to 1,152 chips together into a larger processing system, see Fig.~\ref{fig:loihi_systems} in Appendix \ref{app:loihi_detailed_section}. 
Each neurocore stores weights and recurrent states in its local SRAM, which minimizes memory movement and consequently improves both energy efficiency and latency.

\begin{table*}[h]
\caption{Quantization results for the 370M model, using PyTorch on GPU. All weight matrices are ternary. W8: using 8-bit element-wise weights. A8/A16: using 8-bit or 16-bit activations.
% Ax / Wx: activations / RMSNorm weights quantized to x-bit integers using symmetric per-tensor quantization with power-of-two scales (p2). $\epsilon_\mathrm{rms} \uparrow$: setting the value for $\epsilon_\mathrm{rms}$ to $10^{-3}$ (instead of $\epsilon_\mathrm{rms}=10^{-6}$ as used for training). The last row shows the model configuration used for Loihi 2.
}
\centering
\begin{tabular}{l|ccccccc}
\toprule
 & \textbf{ARCc} & \textbf{ARCe} & \textbf{HS} & \textbf{OQ} & \textbf{PQ} & \textbf{WGe} & \textbf{Avg.} \\
\midrule
Transformer++ & 24.0\% & 45.0\% & 34.3\% & 29.2\% & 64.0\% & 49.9\% & 41.1\% \\
\textbf{Ours} & 23.8\% & 42.6\% & 32.8\% & 28.4\% & 63.0\% & 49.2\% & 40.3\% \\
\midrule
W8A8 & 28.3\% & 26.8\% & 26.1\% & 27.0\% & 52.7\% & 51.5\% & 35.4\% \\
W8A16 & 23.0\% & 42.4\% & 32.4\% & 27.8\% & 63.0\% & 50.1\% & 39.8\% \\
\bottomrule
\end{tabular}
\label{tab:pytorch-mismatch-results}
\end{table*}

To accommodate Loihi 2's low-precision fixed-point arithmetic, we studied the zero-shot accuracy of our 370M-parameter model under various quantization schemes (Table~\ref{tab:pytorch-mismatch-results}). Quantizing normalization parameters and activations to 8-bit weights and 16-bit activations (W8A16) preserves nearly the same performance as the original MatMul-free LM, whereas 8-bit activations (W8A8) reduce accuracy more noticeably, thus activations are kept in 16-bit precision.
All quantized operators were implemented end-to-end in fixed-point arithmetic on Loihi 2, including a custom fixed-point RMSNorm layer (Appendix \ref{app:fxp-implementation}).
% Verification of the model on Loihi 2 indicates close alignment with the quantized PyTorch simulation.
%Inference latencies and for a single block running on a single Loihi 2 chip are characterized in Section \ref{ss:loihi-methods}.

\begin{table*}[h]
\caption{Throughput and energy efficiency for various transformer-based language models with at most 500M parameters running on the NVIDIA Jetson Orin Nano compared to our MatMul-free LM running on Intel's Loihi 2. Bolded metrics are based on single-chip metrics and inter-chip communication results$^\dagger$.}
\centering
\resizebox{\textwidth}{!}{
\begin{tabular}{l|llll|llll}
\toprule
 & \multicolumn{4}{l}{Throughput (tokens/sec)} & \multicolumn{4}{l}{Efficiency (mJ/token)} \\
\midrule
Generate & 500 & 1000 & 2000 & 4000 & 500 & 1000 & 2000 & 4000 \\
\midrule
Alireo-400M & 14.3 & 14.9 & 15.0 & 14.7 & 723 & 719 & 751 & 853 \\
Qwen2-500M & 13.4 & 14.0 & 14.1 & 14.1 & 791 & 785 & 816 & 912 \\
Ours (370M, 1-chip$^*$) & 71.3 & 71.3 & 71.3 & 71.3 & 59 & 59 & 59 & 59 \\
\textbf{Ours (370M, system$^\dagger$)} & \textbf{59.4} & \textbf{59.4} & \textbf{59.4} & \textbf{59.4} & \textbf{70.8} & \textbf{70.8} & \textbf{70.8} & \textbf{70.8} \\
\midrule
Prefill & 500 & 1000 & 2000 & 4000 & 500 & 1000 & 2000 & 4000 \\
\midrule
Alireo-400M & 849.4 & 1620 & 2858 & 3153 & 11.7 & 7.8 & 5.8 & 6.8 \\
Qwen2-500M & 627 & 909 & 1514 & 2639 & 17.9 & 13.9 & 9.5 & 6.7 \\
Ours (370M, 1-chip$^*$) & 13965 & 13965 & 13965 & 13965 & 2.8 & 2.8 & 2.8 & 2.8 \\
\textbf{Ours (370M, system$^\dagger$)} & \textbf{11637} & \textbf{11637} & \textbf{11637} & \textbf{11637} & \textbf{3.4} & \textbf{3.4} & \textbf{3.4} & \textbf{3.4} \\
\bottomrule
\multicolumn{9}{p{12.5cm}}{
\tiny$^*$ The MatMul-free LM on Loihi 2 was characterized on an Oheo Gulch single-chip Loihi 2 system (N3C1 silicon) running NxKernel v0.2.0 and NxCore v2.5.8 (only accessible to Intel Neuromorphic Research Community members). The 1-chip case neglects inter-chip communication.
\par
$^\dagger$ Inter-chip communication causes a derived $\approx$20\% slowdown over the single chip case (Appendix \ref{app:l2-results-detailed}).
% $^\dagger$ Estimates based on single-chip results with 20\% slowdown from inter-chip communication, see Appendix \ref{app:l2-results-detailed}.
\par
$^\ddagger$ Transformer LMs characterized on NVIDIA Jetson Orin Nano 8GB using MAXN power mode running Jetpack 6.2, TensorRT 10.3.0, CUDA 12.4. Energy values include CPU\_GPU\_CV, SOC, and IO components as reported by jtop 4.3.0.
\par
Performance results as of Jan 2025 and may not reflect all publicly available security updates. Results may vary.
}
\end{tabular}
}
\label{tab:jetson-loihi-comp}
\end{table*}

Throughput is at least 4$\times$ higher for autoregressive generation and at least 3.6$\times$ higher for prefill. More details can be found in Appendix~\ref{app:l2-results-detailed}.
Table~\ref{tab:jetson-loihi-comp} contrasts the performance of the 370M MatMul-free model on Loihi~2 against Transformer baselines running on an NVIDIA Jetson Orin Nano, showing at least 1.7$\times$ less energy per token for prefill, and at least 10$\times$ less energy per token for auto-regressive generation. We compare against the 500M parameter Qwen2 model \cite{yang2024qwen2technicalreport}, and also against the 400M parameter Alireo model \cite{alireo2024} which is representative of a small-scale Llama model~\cite{llama3modelcard}.
We have included extended experimental results that compare against server-class GPUs (H100) in the Appendix to provide additional context (\ref{app:detailed-hw-results}), though omit them here due to the significant differences between memory and power draw.

% \section{Discussion}
% \textcolor{red}{TBD.}
\section{Methods}\label{sec:methods}

This section first introduces MatMul-free BitLinear layers with ternary weights (${-1,0,+1}$), replacing multiplications with additions and negations to cut compute and memory while preserving expressiveness (Sec.~\ref{sec:matmul-dense}). It then presents a hardware-efficient fused BitLinear implementation (Sec.~\ref{sec:fused}) and assembles the full \net{} architecture (Sec.~\ref{sec:arch}): a MatMul-free token mixer built on the Linear Gated Recurrent Unit (\name{}) to capture sequence dependencies and a channel mixer that employs a BitLinear-based Gated Linear Unit, so the entire model relies solely on additions and element-wise products. Training details are summarized in Sec.~\ref{sec:training}.

\subsection{MatMul-free Dense Layers with Ternary Weights}
\label{sec:matmul-dense}
In a standard dense layer, the MatMul between the input $x \in \mathbb{R}^{d}$ and the weight matrix $W \in \mathbb{R}^{d \times m}$ can be expressed as:
\begin{equation*}
y = xW = \sum_{j=1}^{d} x_j W_{ij} \quad \text{for } i = 1, 2, \ldots, m
\end{equation*}
where $y \in \mathbb{R}^{m}$ is the output. To avoid using standard MatMul-based dense layers, we adopt BitNet to replace dense layers containing MatMuls with BitLinear modules, which use ternary weights to transform MatMul operations into pure addition operation with accumulation, i.e., ternary accumulation. When using ternary weights, the elements from the weight matrix $W$ are constrained to values from the set $\{-1, 0, +1\}$. Let $\widetilde{\mathbf{W}}$ denote the ternary weight matrix. The MatMul with ternary weights can be expressed as:
\begin{equation*}
\widetilde{\mathbf{Y}} = x \circledast \widetilde{\mathbf{W}} = \sum_{j=1}^{d} x_j \widetilde{\mathbf{W}}_{ij}, \quad \widetilde{\mathbf{W}}_{ij} \in \{-1, 0, +1\}, \quad \text{for } i = 1, 2, \ldots, m
\end{equation*}
where $\widetilde{\mathbf{Y}} \in \mathbb{R}^{m}$ is the output, and $\circledast$ represents a ternary MatMul, which can be simplified to accumulation.
Since the ternary weights $\widetilde{\mathbf{W}}_{ij}$ can only take values from $\{-1, 0, +1\}$, the multiplication operation in the MatMul can be replaced by a simple addition or subtraction operation:
\begin{equation*}
x_j \widetilde{\mathbf{W}}_{ij}  = \begin{cases}
x_j, & \text{if } \widetilde{\mathbf{W}}_{ij} = 1, \\
0, & \text{if } \widetilde{\mathbf{W}}_{ij} = 0, \\
-x_j, & \text{if } \widetilde{\mathbf{W}}_{ij} = -1.
\end{cases}
\end{equation*}
Therefore, ternary MatMul can be written as follows:
\begin{equation*}
\widetilde{\mathbf{Y}}_i = \sum_{j=1}^{d} x_j \widetilde{\mathbf{W}}_{ij} = \sum_{j : \widetilde{\mathbf{W}}_{ij} = 1}x_j-\sum_{j : \widetilde{\mathbf{W}}_{ij} = -1}x_j, \quad \text{for } i = 1, 2, \ldots, m
\end{equation*}
\\
\\
\subsection{Hardware-efficient Fused BitLinear Layer}
\label{sec:fused}
\begin{algorithm*}[t!]
  \scriptsize
  \begin{multicols}{2}
    \newcommand{\round}[1]{\ensuremath{\lfloor#1\rceil}}
    \begin{algorithmic}[1]
    
      \Def \textsc{FordwardPass}($\mathbf{X}, \mathbf{W}, \boldsymbol{b}, \epsilon$)
      \State $\mathbf{X} \in \mathbb{R}^{M \times N}$, $\mathbf{W} \in \mathbb{R}^{N \times K}$, $\boldsymbol{b} \in \mathbb{R}^{K}$
      % \State $\epsilon \leftarrow 1e-6$
      
      \Statex
      \Function{$\mathtt{forward\_pass}$}{$\mathbf{X}, \mathbf{W}, \boldsymbol{b}, \epsilon$}
      \State Load $\mathbf{X}, \mathbf{W}, \boldsymbol{b}, \epsilon$ from HBM
      \State On Chip: $\widetilde{\mathbf{Y}},\mu,\sigma^2,r \gets \mathtt{rms\_norm\_fwd}(\mathbf{X})$
      \State On Chip: $\widetilde{\mathbf{W}} \gets \mathtt{weight\_quant}(\mathbf{W})$
      \State On Chip: $\mathbf{O} \gets \widetilde{\mathbf{Y}} \circledast \widetilde{\mathbf{W}} + \boldsymbol{b}$
      \State Store $\mathbf{O}, \mu,\sigma^2, r$ to HBM
      \State \Return $\mathbf{O}, \mu,\sigma^2, r$
      \EndFunction

      \Statex
      \Function{$\mathtt{rms\_norm\_fwd}$}{$\mathbf{X}$}
      % \State $\mu \gets \mathtt{mean}(\mathbf{X}, \text{axis}=1)$
      % \State $\sigma^2 \gets \mathtt{variance}(\mathbf{X}, \mu)$
      % \State $r \gets 1 / \mathtt{sqrt}(\sigma^2 + \epsilon)$
      
      \State $\mu,\sigma^2 \gets \mathtt{mean}(\mathbf{X}),\mathtt{variance}(\mathbf{X})$
      \State $r \gets \frac{1}{\sqrt{\sigma^2 + \epsilon}}$
      \State $\widetilde{\mathbf{Y}} \gets \mathtt{activation\_quant}(r(\mathbf{X} - \mu) )$
      \State \Return $\widetilde{\mathbf{Y}}, \mu,\sigma^2, r$
      \EndFunction

      \Statex
      \Function{$\mathtt{activation\_quant}$}{$\mathbf{X}$}
      \State $s \gets \frac{127}{\mathtt{max}(|\mathbf{X}|)} $\Comment{$\round{\cdot}\mid_{\cdot}$ means $\mathtt{round}$ then $\mathtt{clamp}$}
      \State $\widetilde{X} \gets \round{s\mathbf{X}}\mid_{[-128, 127]}\cdot \frac{1}{s} $
      \State \Return $\widetilde{X}$
      \EndFunction

      \Statex
      \Function{$\mathtt{weight\_quant}$}{$\mathbf{W}$}
      \State $s \gets \frac{1}{\mathtt{mean}(|\mathbf{W}|)}$
      \State $\widetilde{\mathbf{W}} \gets \round{s\mathbf{X}}\mid_{[-1,1]} \cdot \frac{1}{s}$
      \State \Return $\widetilde{\mathbf{W}}$
      \EndFunction

      \State \Return $\mathbf{O}$

    \end{algorithmic}

    \columnbreak

    \begin{algorithmic}[2]
      \Def \textsc{BackwardPass}($\mathbf{X}, \mathbf{W}, \boldsymbol{b}, \mathbf{O}, \mathrm{d}\mathbf{O}, \mu, \sigma^2, r$)
      \State $\mathbf{X} \in \mathbb{R}^{M \times N}$, $\mathbf{W} \in \mathbb{R}^{N \times K}$, $\boldsymbol{b} \in \mathbb{R}^{K}$
      \State $\mathbf{O} \in \mathbb{R}^{M \times K}$, $\mathrm{d}\mathbf{O} \in \mathbb{R}^{M \times K}$

      \Statex
      \Function{$\mathtt{backward\_pass}$}{$\mathbf{X}, \mathbf{W}, \boldsymbol{b}, \mathbf{O}, \mu, \sigma^2, r, \mathrm{d}\mathbf{O}$}
      \State Load $\mathbf{X}, \mathbf{W}, \boldsymbol{b}, \mathbf{O}, \mu, \sigma^2, r, \mathrm{d}\mathbf{O}$ from HBM
      \State On Chip: $\mathrm{d}\mathbf{Y} \gets \mathrm{d}\mathbf{O} \times \mathbf{W}^\top$
      \State On Chip: $\mathbf{dX},\widetilde{\mathbf{Y}} \gets \mathtt{rms\_norm\_bwd}(\mathrm{d}\mathbf{Y}, \mathbf{X}, \mu, \sigma^2, r)$
      
      \State On Chip: $\mathrm{d}\mathbf{W} \gets \mathrm{d}\mathbf{O}^\top \times \widetilde{\mathbf{Y}}$
      \State On Chip: $\mathrm{d}\boldsymbol{b} \gets \mathtt{sum}(\mathrm{d}\mathbf{O})$
      \State Store $\mathbf{dX}, \mathrm{d}\mathbf{W}, \mathrm{d}\boldsymbol{b}$ to HBM
      \State \Return $\mathbf{dX}, \mathrm{d}\mathbf{W}, \mathrm{d}\boldsymbol{b}$
      \EndFunction

      \Statex
      \Function{$\mathtt{rms\_norm\_bwd}$}{$\mathrm{d}\mathbf{Y}, \mathbf{X}, \mu, \sigma^2, r$}
      \State $\widetilde{\mathbf{Y}} \gets \mathtt{activation\_quant}(r(\mathbf{X} - \mu) )$
      \State $\mathrm{d}\sigma^2 \gets \mathtt{sum}(\mathrm{d}\mathbf{Y} \times (\mathbf{X} - \mu) \times -0.5 \times r^3)$
      \State $\mathrm{d}\mu \gets \mathtt{sum}(-r\mathrm{d}\mathbf{Y}) + \mathrm{d}\sigma^2 \times \mathtt{mean}(\mathbf{X} - \mu)$
      \State $\mathbf{dX} \gets r\mathrm{d}\mathbf{Y}  + 2\mathrm{d}\sigma^2 (\mathbf{X} - \mu) / N + \mathrm{d}\mu / N$
      \State \Return $\mathbf{dX},\widetilde{\mathbf{Y}}$
      \EndFunction

    \end{algorithmic}
  \end{multicols}
  \caption{Fused RMSNorm and BitLinear Algorithm with Quantization}
  \vspace{-10pt}
  \label{alg:fused_rms_linear_quant}
  
\end{algorithm*}

BitNet showed that stabilizing ternary layers requires an additional RMSNorm before the BitLinear input. However, the vanilla implementation of BitNet is not efficient. Modern GPUs feature a memory hierarchy with a large, global high-bandwidth memory (HBM) and smaller, faster shared memory (SRAM), and the implementation of BitNet introduced many I/O operations: reading the previous activation into SRAM for RMSNorm, writing it back for quantization, reading it again for quantization, storing it, and reading it once more for the Linear operation. To address this inefficiency, we read the activation only once and apply RMSNorm and quantization as fused operations in SRAM, which we present in Algorithm~\ref{alg:fused_rms_linear_quant}.
% presents our approach for improving the hardware efficiency of the BitLinear layer by fusing quantized RMSNorm and BitLinear operations. 
Optimal utilization of SRAM to reduce HBM I/O costs can significantly speed up computations. Since the activations in this model have a larger memory footprint than ternary weights and the large number of element-wise operations, our optimization efforts focus on activations.

\begin{wrapfigure}{r}{0.5\textwidth}
  \begin{center}
    \includegraphics[width=0.48\textwidth]{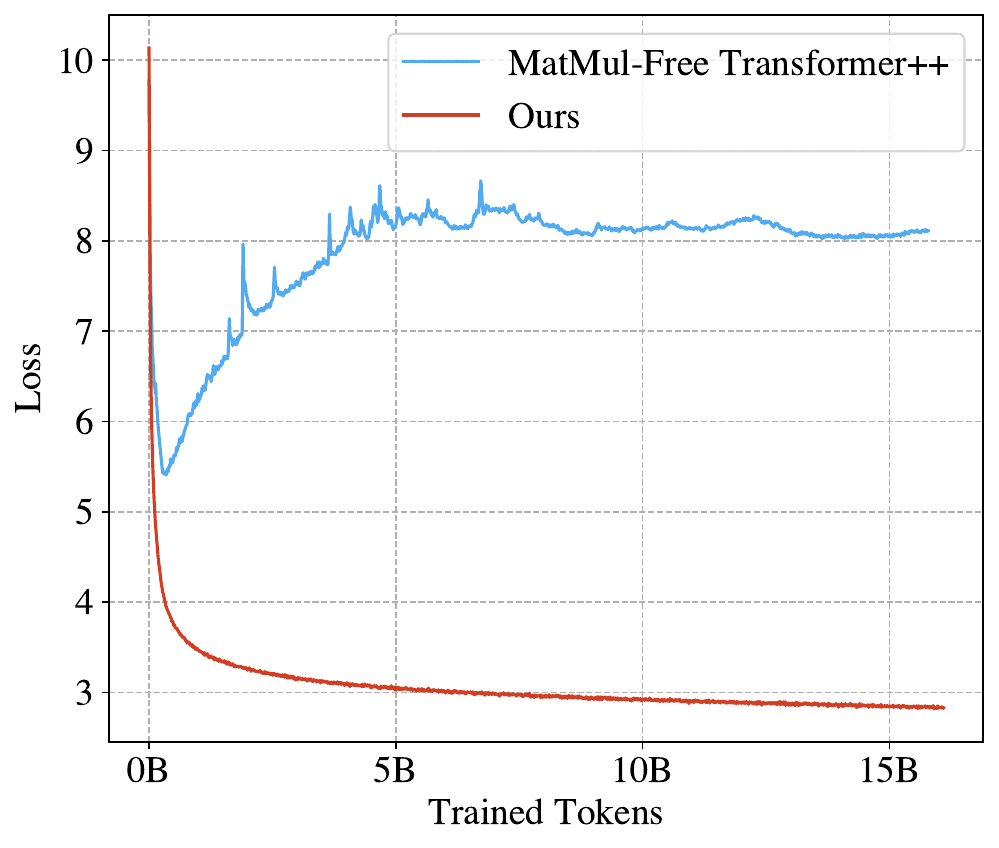}
  \end{center}
  \caption{Training loss over steps for the MatMul-free Transformer++ and our proposed method in 370M. The MatMul-free Transformer++ fails to converge, while our method successfully converges under the MatMul-free setting.}
  \label{fig:training_loss}
\end{wrapfigure}

The \texttt{forward\_pass} function in Algorithm~\ref{alg:fused_rms_linear_quant} first calls \texttt{rms\_norm\_fwd} to perform RMSNorm on input activations $\mathbf{X}$, loading normalized activations $\mathbf{Y}$, mean $\mu$, variance $\sigma^2$, and scaling factor $r$ from HBM. The normalized activations $\mathbf{Y}$ are then quantized to obtain $\widetilde{\mathbf{Y}}$ and the weights $\mathbf{W}$ are quantized using \texttt{weight\_quant}, with both performed without off-chip data movement. Finally, the output $\mathbf{O}$ is computed on-chip by multiplying quantized activations $\widetilde{\mathbf{Y}}$ with the ternary weights $\widetilde{\mathbf{W}}$, adding the bias $\boldsymbol{b}$, and then storing the result back to HBM.

The \texttt{backward\_pass} function first loads $\mathbf{X}$, $\mathbf{W}$, $\boldsymbol{b}$, $\mathbf{O}$, $\mu$, $\sigma^2$, $r$, and the output gradient $\mathrm{d}\mathbf{O}$ from HBM. The gradient $\mathrm{d}\mathbf{Y}$ is then computed on-chip by multiplying the output gradient $\mathrm{d}\mathbf{O}$ with the transposed weight matrix $\mathbf{W}^\top$. Next, it calls \texttt{rms\_norm\_bwd} on-chip to backpropagate through RMSNorm, computing the input gradient $\mathrm{d}\mathbf{X}$. The weight gradient $\mathrm{d}\mathbf{W}$ is calculated on-chip by multiplying the transposed output gradient $\mathrm{d}\mathbf{O}^\top$ with the quantized activations $\widetilde{\mathbf{Y}}$, and the bias gradient $\mathrm{d}\boldsymbol{b}$ is obtained by summing $\mathrm{d}\mathbf{O}$. The computed gradients $\mathbf{dX}$, $\mathrm{d}\mathbf{W}$, and $\mathrm{d}\boldsymbol{b}$ are then transferred back to HBM. Sec.~\ref{sec:exp-fused} presents an experimental comparison between a vanilla BitLinear implementation and Fused BitLinear.

\subsection{MatMul-free Language Model Architecture}
\label{sec:arch}
We adopt the perspective from Metaformer~\cite{yu2022metaformer}, which suggests that Transformers consist of a token-mixer (for mixing temporal information, i.e., Self Attention~\cite{vaswani2017attention}, Mamba~\cite{gu2023mamba}) and a channel-mixer (for mixing embedding/spatial information, i.e., feed-forward network, Gated Linear Unit (GLU) ~\cite{dauphin2017glu, shazeer2020glu}). A high-level overview of the architecture is shown in Fig.~\ref{fig:main}. 
 
\subsubsection{MatMul-free Token Mixer}

Self-attention is the most common token mixer in modern language models, relying on matrix multiplication between three matrices: $Q$, $K$, and $V$. To convert these operations into additions, we ternarize at least two of the matrices. Assuming all dense layer weights are ternary, we quantize $Q$ and $K$, resulting in a ternary attention map that eliminates multiplications in self-attention.
However, as shown in Fig.~\ref{fig:training_loss}, such a model fails to converge. One possible explanation is that activations contain outliers crucial for performance but difficult to quantize effectively~\cite{pan2023smoothquant+, xiao2023smoothquant}. To address this challenge, we explore alternative methods for mixing tokens without relying on matrix multiplications.

By resorting to the use of ternary RNNs, which combine element-wise operations and accumulation, it becomes possible to construct a MatMul-free token mixer.
Among various RNN architectures, the GRU is noted for its simplicity and efficiency, achieving similar performance to Long Short-Term Memory (LSTM)~\cite{hochreiter1997long} cells while using fewer gates and having a simpler structure. Thus, we choose the GRU as the foundation for building a MatMul-free token mixer. We first revisit the standard GRU and then demonstrate, step by step, how we derive the \name{}.

\paragraph{Revisiting the Gated Recurrent Unit} The GRU~\cite{cho2014learning} 
% is a widely-used variant of the RNN architecture that is simpler and more computationally efficient compared to the LSTM unit while still maintaining comparable performance. 
can be formalized as follows:
\begin{align}
    \boldsymbol{r}_t &= \sigma\left(\boldsymbol{x}_t \mathbf{W}_{xr} + \boldsymbol{h}_{t-1} \mathbf{W}_{hr} + \mathbf{b}_{r}\right) \in \mathbb{R}^{d}, \\
    \boldsymbol{f}_t &= \sigma\left(\boldsymbol{x}_t \mathbf{W}_{xf} + \boldsymbol{h}_{t-1} \mathbf{W}_{hf} + \mathbf{b}_{f}\right) \in \mathbb{R}^{d}, \\
    \boldsymbol{c}_t &= \tanh\left(\boldsymbol{x}_t \mathbf{W}_{xc} + (\boldsymbol{r}_t \odot \boldsymbol{h}_{t-1}) \mathbf{W}_{cc} + \mathbf{b}_{c}\right) \in \mathbb{R}^{d}, \\
    \label{eq:tied}
    \boldsymbol{h}_t &= \boldsymbol{f}_t \odot \boldsymbol{h}_{t-1} + (1 - \boldsymbol{f}_t) \odot \boldsymbol{c}_t \in \mathbb{R}^{d}, \\
    \boldsymbol{o}_t &= \boldsymbol{h}_t
\end{align}
where $\boldsymbol{x}_t \in \mathbb{R}^{m}$ is the input vector at time step $t$, $\boldsymbol{h}_{t-1} \in \mathbb{R}^{d}$ is the hidden state vector from the previous time step, $\boldsymbol{r}_t \in \mathbb{R}^{d}$ is the reset gate vector, $\boldsymbol{f}_t \in \mathbb{R}^{d}$ is the forget gate vector, $\boldsymbol{c}_t \in \mathbb{R}^{d}$ is the candidate hidden state, $\boldsymbol{h}_t \in \mathbb{R}^{d}$ is the final hidden state vector at time step $t$, $\boldsymbol{o}_t \in \mathbb{R}^{d}$ is the output vector at time step $t$, $\mathbf{W}{(\cdot)} \in \mathbb{R}^{m \times d}$ and $\mathbf{b}{(\cdot)} \in \mathbb{R}^{d}$ are learnable weight matrices and bias vectors, respectively, $\sigma(\cdot)$ is the sigmoid activation function, and $\odot$ denotes element-wise multiplication. 

A key characteristic of the GRU is the coupling of the input gate vector $\boldsymbol{f}_t$ and the forget gate vector $(1 - \boldsymbol{f}_t)$, which together constitute the `leakage' unit. This leakage unit decays the hidden state $\boldsymbol{h}_{t-1}$ and the candidate hidden state $\boldsymbol{c}_t$ through element-wise multiplication (Eq.~\ref{eq:tied}). This operation allows the model to adaptively retain information from the previous hidden state $\boldsymbol{h}_{t-1}$ and incorporate new information from the candidate hidden state $\boldsymbol{c}_t$. Importantly, this operation relies solely on element-wise multiplication, avoiding the need for MatMul. We preserve this property of the GRU while introducing further modifications to create a MatMul-free variant of the model.

\paragraph{MatMul-free Linear Gated Recurrent Unit} We first remove hidden-state related weights $\mathbf{W}_{cc}$, $\mathbf{W}_{hr}$, $\mathbf{W}_{hf}$, and the activation between hidden states ($\tanh$). This modification not only makes the model MatMul-free but also linearized the GRU through time, enabling training via a parallel scan~\cite{gu2023mamba, voelker2019legendre}. This approach is critical for improving computational efficiency, as transcendental functions are expensive to compute accurately, and non-diagonal transition matrices increase runtime complexity from $\mathcal{O(n)}$ to $\mathcal{O(n^3)}$. This modification is a key feature of recent RNNs, such as the Linear Recurrent Unit~\cite{orvieto2023lru}, Hawk~\cite{de2024griffin}, and RWKV-4~\cite{rwkv4}. We then add a data-dependent output gate between $\boldsymbol{h}_t$ and $\boldsymbol{o}_t$, inspired by the LSTM and widely adopted by recent RNN models:
\begin{equation*}
\begin{aligned}
        \boldsymbol{g}_t &= \boldsymbol{x}_t \mathbf W_g  + \mathbf b_g  \in \mathbb R^{ d},  \\
\boldsymbol{o}^{\prime}_t &= \boldsymbol{g}_t \odot \sigma (\boldsymbol{h}_t) \in \mathbb R^{ d}, \\
\boldsymbol{o}_t  &=   \boldsymbol{o}^{\prime}_t \mathbf W_{o} + \mathbf b_o \in \mathbb R^{ d}.
\end{aligned}
\end{equation*}
Following the approach of HGRN~\cite{qin2024hgrn}, we further simplify the computation of the candidate hidden state by keeping it as a simple linear transform, rather than coupling it with the hidden state. This can be rewritten as a linear transformation of the input. Finally, we replace all remaining weight matrices with ternary weight matrices, completely removing the MatMul operations.  
The resulting \name{} architecture can be formalized as follows:
\begin{equation*}
\begin{aligned}
     \boldsymbol{f}_t &=\sigma\left(\boldsymbol{x}_t \circledast \mathbf W_{f}+ \mathbf b_{f}\right) \in \mathbb R^{ d},\\
    % \lambda_{t}&= \gamma^k + (1 - \gamma^k) \odot \mu_t  \in \mathbb R^{ d},\\
\boldsymbol{c}_t & =\tau\left(\boldsymbol  x_t \circledast \mathbf W_c + \mathbf b_c\right) \in \mathbb R^{ d},\\
\boldsymbol{h}_t & = \boldsymbol{f}_t \odot \boldsymbol{h}_{t-1}+ (1-\boldsymbol{f}_t) \odot \boldsymbol{c}_t \in \mathbb R^{ d},\\
\boldsymbol{g}_t &= \boldsymbol{x}_t \circledast \mathbf W_g  + \mathbf b_g  \in \mathbb R^{ d},  \\
\boldsymbol{o}^{\prime}_t &= \boldsymbol{g}_t \odot \sigma (\boldsymbol{h}_t) \in \mathbb R^{ d}, \\
\boldsymbol{o}_t  &=   \boldsymbol{o}^{\prime}_t \circledast \mathbf W_{o} + \mathbf b_o \in \mathbb R^{ d}.
\end{aligned}
\end{equation*}
where $\mathbf W_c, \mathbf W_{f},\mathbf W_o,\mathbf W_g \in \mathbb{R}^{d \times d}$ consist of ternary weights, $\boldsymbol{f}_t$ is the forget gate output, $\sigma$ is the Sigmoid activation function,  $\boldsymbol{c}_t$ is the input vector, $\tau$ is the SiLU activation function, $\boldsymbol{h}_t$ is the hidden state, $\boldsymbol{g}_t$ is the output gate, $\boldsymbol{o}^{\prime}_t$ is the intermediate output, and $\boldsymbol{o}_t$ is the final output at time step $t$. The initial hidden state $\boldsymbol{h}_0$ is set to $\mathbf{0}$. 
Similarly to HGRN, we also employ the $\mathrm{cummax}$ operation to bound the forget gate values in deeper layers closer to 1, though omit this above for brevity. The \name{} can be viewed as a simplified variant of HGRN that omits complex-valued components and reduces the hidden state dimension from $2d$ to $d$. This simplification makes \name{} more computationally efficient while preserving essential gating mechanisms and ternary weight quantization.

Recurrent LLMs are known to have limitations in long-context benchmarks and retrieval. Linear recurrent models partially address this by 1) avoiding non-linearities through time to improve gradient flow, 2) introducing data-dependent decay, and 3) introducing lower-bounds on the `forget gate'~\cite{qin2024hgrn}. \net{} adopts these best practices, and recent work shows that hybrid architectures with very few Transformer blocks can compensate for RNN performance across longer contexts~\cite{wang2025systematic}.

% Alternatively to the MLGRU, which employs a data-dependent decay with element-wise product hidden state, the a similarly modified version of the RWKV-4 model can also satisfy the requirement of a MatMul-free token mixer, utilizing static decay and normalization. The performance of using RWKV-4 as a MatMul-free token mixer is discussed in the Experiment section, with a detailed description of the RWKV-4 model provided in Appendix~\ref{appendix:rwkv4}. However, RWKV-4 introduces exponential and division operations, which are less hardware-efficient compared to the MLGRU.
 
\subsubsection{MatMul-free Channel Mixer}

For MatMul-free channel mixing, we use GLU, which is widely adopted in many modern LLMs, including Llama~\cite{touvron2023llama,touvron2023llama2, llama3modelcard}, Mistral~\cite{jiang2023mistral} and RWKV~\cite{rwkv4}, and a BitLinear-adapted version can be expressed as follows:
\begin{equation*}
  \begin{aligned}
\boldsymbol{g}_t &= \boldsymbol{x}_{t}\circledast \boldsymbol{W}_g \in \mathbb R^{ l}, \\
\boldsymbol{u}_t &= \boldsymbol{x}_{t}\circledast \mathbf{W}_u\in \mathbb R^{ l}, \\
\boldsymbol{p}_t &= \tau(\boldsymbol{g}_t) \odot \boldsymbol{u}_t\in \mathbb R^{ l}, \\
\boldsymbol{d}_t &= \boldsymbol{p}_t\circledast \mathbf{W}_d\in \mathbb R^{ d}, 
\end{aligned}  
\end{equation*}
where $\tau$ denotes the SiLU activation function, $\circledast$ represents ternary accumulation, and $\odot$ represents the element-wise product.

The GLU consists of three main steps: 1) \emph{upscaling} the $t$-step input $\boldsymbol{x}_t \in \mathbb{R}^{d}$ to $\boldsymbol{g}_t,\boldsymbol{u}_t\in\mathbb{R}^{l}$ using weight matrices $\mathbf{W}_g, \mathbf{W}_u \in \mathbb{R}^{d \times l}$ 2) \emph{elementwise gating} $\boldsymbol{u}_t$ with $\boldsymbol{g}_t$ followed by a nonlinearity $f(\cdot)$, where we apply $\mathrm{Swish}$ \cite{shazeer2020glu}. 3) \emph{Down-scaling} the gated representation $\boldsymbol{p}_t$ back to the original size through a linear transformation $\mathbf{W}_d$. 
Following Llama \cite{touvron2023llama}, we maintain the overall number of parameters of GLU at $8 d^2$ by setting the upscaling factor to $\frac{8}{3}d$.

The channel mixer here only consists of dense layers, which are replaced with ternary accumulation operations. By using ternary weights in the BitLinear modules, we can eliminate the need for expensive MatMuls, making the channel mixer more computationally efficient while maintaining its effectiveness in mixing information across channels.

\subsection{Training Details}
\label{sec:training}
\paragraph{Surrogate Gradient}
To handle non-differentiable functions such as the Sign and Clip functions during backpropagation, we use the straight-through estimator (STE) \cite{ste} as a surrogate function for the gradient. STE allows gradients to flow through the network unaffected by these non-differentiable functions, enabling the training of our quantized model. This technique is widely adopted in BNNs and SNNs.
\paragraph{Large Learning Rate}
When training a language model with ternary weights, using the same learning rate as regular models can lead to excessively small updates that have no impact on the clipping operation. This prevents weights from being effectively updated and results in biased gradients and update estimates based on the ternary weights. To address this challenge, it is common practice to employ a larger learning rate when training binary or ternary weight language models, as it facilitates faster convergence~\cite{zhang2024binarized, liu2023binary, ma20241bitllm}.
In our experiments, we maintain consistent learning rates across both the 370M and 1.3B models, aligning with the approach described in Ref.~\cite{qin2023scaling}. Specifically, for the Transformer++ model, we use a learning rate of $3e-4$, while for the \net{}, we employ a learning rate of $4e-3$, $2.5e-3$, $1.5e-3$ in 370M, 1.5B and 2.7B, respectively. These learning rates are chosen based on the most effective hyperparameter sweeps for faster convergence during the training process.

\paragraph{Learning Rate Scheduler}
When training conventional Transformers, it is common practice to employ a cosine learning rate scheduler and set a minimal learning rate, typically $0.1 \times$ the initial learning rate. We follow this approach when training the full precision Transformer++ model. However, for the \net{}, the learning dynamics differ from those of conventional Transformer language models, necessitating a different learning strategy. We begin by maintaining the cosine learning rate scheduler and then reduce the learning rate by half midway through the training process. Interestingly, we observed that during the final training stage, when the network's learning rate approaches 0, the loss decreases significantly, exhibiting an \textit{S}-shaped loss curve. This phenomenon has also been reported by ~\cite{ma20241bitllm, zhang2024binarized} when training binary/ternary language models.

% \section{Experiments}
% \subsection{Learning Rate Analysis}
% The learning rate is a crucial hyper-parameter in language model training, and models become more sensitive to the learning rate in the ternary/binary weight regime. To determine the optimal learning rate, we conducted a search within the range of $1.5e-3$ to $3e-2$ using our 370M model with a batch size of 50k tokens. The results of this search are shown in Fig.~\ref{fig:merge_result}(c).
% The results demonstrate that the final training loss monotonically decreases as the learning rate increases from $1.5e-3$ to $1e-2$. The model only exhibits instability when the learning rate exceeds $2e-2$. This finding suggests that previous works employing ternary weights, such as BitNet, which uses a learning rate of $1.5e-3$, may not be optimal and that higher learning rates could potentially lead to better performance. These findings align with the observations from the Deepseek LLM~\cite{bi2024deepseek} which found that the optimal learning rate for conventional LLMs is actually larger than the values typically reported in most LLM training setups. Interestingly, we also observed that models trained with larger learning rates at the start of the training process exhibit a more rapid decrease in training loss during the later stages of training compared to those trained with smaller learning rates.

\subsection{Implementation Details on Intel Loihi 2}
\label{ss:loihi-methods}

% The present section describes details for implementing the MatMul-free LM on the Loihi 2 chip from Intel, expanding on the results reported in 
In order to run the MatMul-free LM on the Intel Loihi 2 chip, the original model must be quantized, implemented using fixed-point arithmetic using custom microcode on the Loihi 2 platform, and finally verified against the original model to make sure that the mismatch between the fixed-point model on Loihi 2 and the original model on GPU is limited. We present these steps in the following sections.

\paragraph{Model Quantization}

As a first step, we quantize all weights and activations of the MatMul-free LM and verify the resulting performance of the quantized model on the downstream tasks presented previously. All weight matrices are already ternarized, thus not needing further quantization. However, every BitLinear layer with a ternary weight matrix $W \in \mathbb{R}^{d \times m}$ is preceded by a root mean square normalization (RMSNorm) operation, given by:
\begin{equation}
    \text{RMSNorm}(x; w, \epsilon) = \frac{x}{\sqrt{\epsilon + \sum_i^d x_i^2}} \odot w
\end{equation}
where $w \in \mathbb{R}^d$ is a learned vector of element-wise scaling factors, $\epsilon$ is a small constant set to $\epsilon=10^{-6}$ to avoid zero division, and $x \in \mathbb{R}^d$ is the input activation vector.
% For the MatMul-free LM to run on the Loihi 2 chip, all operations must be implemented in fixed-point arithmetic
While the original MatMul-free model stores all normalization scales $w \in \mathbb{R}^d$ as 16-bit floating-point numbers, we quantize all normalization scales across the model to 8-bit integers using $w_q = \text{round} \left( w / s_w \right)$ where $s_w$ is the quantization scale, chosen as $s_w = 2^{\lfloor \log_2(\max(|w|)) \rceil}$ where $\lfloor \cdot \rceil$ denotes the rounding operation, and $\max(|x|)$ is the absolute maximum activation observed on the data across all activation channels. These quantization scales allow the rescaling between different quantization scales to be efficiently implemented on Loihi 2 using bit-shift operations. Quantizing all weights in this scheme results in a relative performance drop of 0.18\% averages across all downstream tasks reported in Section \ref{sec:performance-downstream-tasks}. 

Whereas activations of the MatMul-free LM are stored as 16-bit floating-point numbers, activations on Loihi 2 are sent between neurons as integer payloads. To test the performance degradation of the model when quantization activations to lower bit precision, the model was quantized using dynamic quantization in PyTorch, where quantization scales for the activations $s_a$ are computed for each batch individually. 
When quantizing activations in the RMSNorm layer, the $\epsilon$ value underflows in most layers, potentially leading to division by zero. We therefore set $\epsilon=10^{-3}$ and report results with this modified value.
Table \ref{tab:pytorch-mismatch-results} shows that for the 370M model, performance drops by less than 1.5\% relative to the original model when using 16-bit activation quantization. Performance drops more significantly, by over 12\% relative to the original model, when using 8-bit activations. All further results will assume 8-bit weights and 16-bit activations. We note that these results are for naive post-training quantization and may be significantly improved through quantization-aware training or fine-tuning the model with specific quantization requirements. 

\paragraph{Fixed-Point Arithmetic}

Once all quantization scales are computed, the model can be run in static quantization with integer weights and activations. However, some operations used in the MatMul-free LM are not defined on integers, namely the sigmoid activation function $\sigma$ and the inverse-square-root in the RMSNorm. 

We employ a look-up-table (LUT) for a fixed-point approximation of the logistic sigmoid function, $\sigma(x) = 1/(1 + e^{-x})$. Specifically, we scale the floating-point input $x$ by $2^{x_{\text{exp}}}$ where $x_{\text{exp}}=6$, quantize it to an integer domain, and store precomputed \(\lfloor \sigma\!\bigl(\frac{x}{2^{x_{\text{exp}}}}\bigr)\cdot 2^{y_{\text{exp}}}\rfloor\) values in a LUT for positive inputs. For negative inputs, we exploit $\sigma(-x) = 1 - \sigma(x)$, thus requiring only half-range values. During inference, a piecewise linear interpolation between adjacent LUT entries refines the output. This design offers efficient computation and controllable approximation error.

For the inverse-square-root in the RMSNorm layer, we adapted a well-known “fast inverse square root” algorithm \texttt{Fast InvSqrt} to operate entirely in fixed-point arithmetic. Our method treats the input $\tilde{x}$ as an integer paired with a fixed exponent, and uses a LUT with 24 values to produce an initial guess for $\sqrt{\tilde{x}}$. This estimate is then refined using five iterations of the Newton-Raphson method, all in a fixed-point format. See Appendix~\ref{app:fxp-implementation} for more details. 

\paragraph{Mapping the MatMul-free LM onto Loihi}

The model on Loihi 2 is implemented as a set of neurons that are connected through synapses. Each neuron asynchronously executes a simple microcode program and sends its output to a collection of other neurons through synapses. Because every neuron contains only information pertaining to its own dynamics, aggregate operations--such as the sum of squares over an activation vectors--must be implemented through an additional neuron that receives inputs from all neurons in the relevant layer. Fig.~\ref{fig:loihi-mmf-graph} in Appendix~\ref{app:jetson-results-detailed} shows the partitioning of the MatMul-free model into Loihi-compatible neurons.

% RMSNorm:
% https://mermaid.ink/pdf/pako:eNp1klFPgzAUhf9Kw163OEeB2gejyXzwwZk4Ex_ELLUUWWQta4vbsuy_W6BEKHjfzjn36720nD0qEuZhL83FgWZEavC6jDlQ5eeXJEUGlP8eey9P65WQu9j7MJEtvjDBIy9K3bf_7YdVf8K43uoTuAKrfhrVp_2s99I5D3WwfhKY5G3DB4NC4z-X-m8zxpMm5gswm92aHa30jQTHjdqXRLKkyWAnMzKwMmhkaCWsyB0j5qa6cGTjqJHISlR1SyH0CGJ2aZe7OwOVkYJhIBnVl3aRcRuO29G4jcbtYNwOR22zIca4EOaFOvpeZa30ndgfxpyVUvDWgQ4A-0Al1YmTQrHWihwi6hPRYARyANQH0HBE4BBBnwiGROgQYZegOVFqyVJQdYB0m-d4giCl4dzJDQGUluKbzQ7bRGf4ujhOWych5j2kJCfMBWetjSc3dU3tuQ91TQEVuZB4Mq_LGdNcjwUYZTRNnQ77dbYlTWkCYfWTxty7_AJFwTxy?fit
% full graph:
% https://mermaid.ink/pdf/pako:eNqtVvGL2jAU_ldChbGBgolFbRkHtzmOgW6H3rEf5pDaphpWk5K0qBz735fW1mti0jm5_lBtvu99730vL6UvTsgi7PhOnLB9uA14Bp4mSwpEvt7wIN2CJDhivur_XDpfaZpnS-eXRKuL4pwzegFiGjUVdsmG55Iymz7Mn03xngTns8U3xncKLI40SAVejcsMn0g2JRQHHMA-cg_F7cZ8EEp0QTY7RiJjQgjfOiMqM06fzenQW6cb1OjHNb97Yr8xBTNyIHRzq6B7sUVXS5xdDv_XZSrp8m4qCJ0cPoLFnjxMS5uftwGlOFGNqiUgfV_RGA6bJWh0dBW97lFjjMH7Hyta_Ma5wNGHWwxe0XFT-Ll2Y7cLA2a10zFHIxnzPc8sB72E51iQKA8ScB9FJCOMmgUFrHbokePexdGuezb6l0uB6qFqF5KsCcvXCQYmOe1sg17v7vVloOFIw9EJr991CgbevbZGoUG35DVnX8U9BS8GU2o1B0-hI1UOneUUgUZlaKDhyIyfHxXQ1YKHWrDmVbWKFCtQdTbWCx-r0p6qrJQFtbKgVhbUyhopoKaMJAgmWI69wlLbAuE1LE_HqumsJ8b3_dNfdd1rWU8Zz6rVRqMkUD2pERDqUs3uW6OQNQq1RA0sZUO3DTAYKrbPlgUNbLUhk6PmlFsLt_Ubag0_HwILH1n8oDY_I1vyVuCyKmijK-wwCYSY4BgUSyAmSeJ3xm4YDvsafi_kGzbj8gOhtydRtvVheujWK1Eg5PcgD44-ZRTXy37HK69upfulvLogZAnjfqdfXlqaU5FVAA5xGMcao-paRYnjMHLd4iAtqfPnLxmRXSs?fit

\clearpage
\bibliography{sn-bibliography}
\bibliographystyle{unsrt}

\clearpage
% \section*{Appendix}
\appendix
\section{FPGA Implementation and Results}
Neither GPUs nor Intel Loihi~2 support 2-bit weights, so we additionally did an RTL-only implementation of the \net{} on a D5005 Stratix 10. 

\subsection{Implementation}
\begin{figure}[h]
    \centering
    \includegraphics[width=1\textwidth]{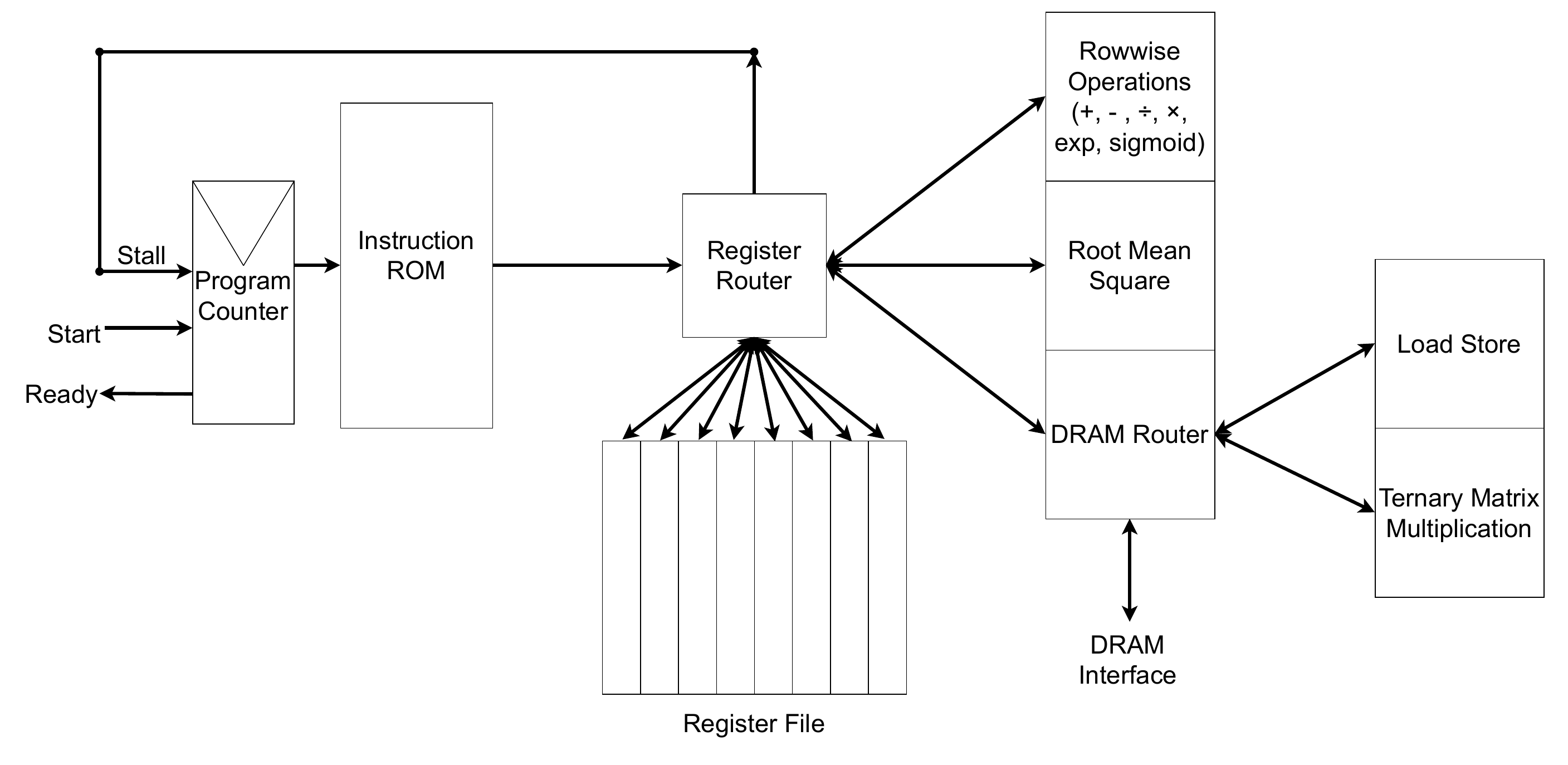}
    \caption{RTL implementation for running MatMul-free token generation}
    \label{fig:rtl diagram}
\end{figure}

To test the power usage and effectiveness of the \net{} on custom hardware that can better exploit ternary operations, we created an FPGA accelerator in SystemVerilog. The overview is shown in Fig.~\ref{fig:rtl diagram}.

There are 4 functional units in this design: \enquote{Row-wise Operation,} \enquote{Root Mean Square,} \enquote{Load Store,} and \enquote{Ternary Matrix Multiplication,} and they each allow for simple out-of-order execution. We wrote a custom assembler for our custom instruction set, which was used to convert assembly files into an instruction ROM. The custom instruction set is given below:

\begin{itemize}
    \item LDV: LoaD Vector from memory
    \item SV: Store Vector to memory
    \item ADD: row-wise ADDition
    \item SUB: row-wise SUBtraction
    \item MUL: row-wise MULtiplication
    \item DIV: row-wise DIVision
    \item EXP: row-wise EXPonential function
    \item SIG: row-wise SIGmoid
    \item NORM: NORMalization with root-mean-square
    \item TMATMUL: Ternary MATrix MULtiplication
\end{itemize}

\paragraph{The register router} delegates incoming instructions to available registers. The register file consists of 8 registers, each storing 1 vector in a separate SRAM array. Each register SRAM array has a read and write port that are delegated to at most one instruction at a time. If an instruction requests access to a functional unit or a register that is busy, the program counter will stall until the functional unit or register has been freed. If two instructions do not block each other, they execute simultaneously.

\paragraph{The \enquote{Root Mean Square} functional unit} uses a specialized hardware algorithm to preserve precision, and runs in 3 stages. Stage 1 will copy the target vector to an internal-temporary register, and perform a square on each element using a lookup-table. Stage 2 will divide-and-conquer to average neighboring vector elements, generating the Root-Mean-Square result. Stage 3 will perform normalization by dividing each element in the original vector by the Root-Mean-Square result. By using divide-and-conquer for averaging, instead of a typical rolling sum then large divide, rounding errors are significantly reduced.

\paragraph{The \enquote{Ternary Matrix Multiplication} functional unit} takes in a DRAM address for a ternary matrix, then performs a TMATMUL on the specified vector. Our architecture entirely places the ternary matrices in DRAM. While running a TMATMUL instruction, an SRAM FIFO is simultaneously filled with sequential DRAM fetch results, and emptied by a power-efficient ternary-add operation. At the moment, the three required TMATMUL instructions take up nearly all of the total execution time. In future work, we will introduce parallelism and caching to improve TMATMUL execution time.

\subsection{Results}
The RTL implementation of the MatMul-free token generation core is deployed on a D5005 Stratix 10 programmable acceleration card (PAC) in the Intel FPGA Devcloud. The core completes a forward-pass of a block in 43ms at $d=512$ and achieves a clock rate of 60MHz. The resource utilization, power and performance of the single-core implementation of a single block ($N=1$) are shown in Tab.~\ref{tab:fpga-metrics1}. `\% ALM Core' refers to the percentage of the total adaptive logic modules used by the core logic, and `\%ALM Total' includes the core, the additional interconnect/arbitration logic, and ``shell'' logic for the FPGA Interface Manager. `M20K' refers to the utilization of the memory blocks, and indicates that the number of cores are constrained by ALMs, and not on-chip memory (for this DDR implementation). We implement a single token generation core, and estimate the total number of cores that could fit on the platform and the corresponding power, performance and area impact. This is the simplest case where the core only receives 8 bits at a time from memory.

\begin{table*}[t]
\setlength{\tabcolsep}{6pt}
\centering
\caption{MatMul-free token generation FPGA core resource utilization and performance metrics. The ternary matrix multiplication operation dominates latency for the current implementation and there is not an observed bottleneck in the local DDR4 bridge. In future implementations, this functional unit will be optimized and the DDR interface will likely become the primary bottleneck.}
\resizebox{\textwidth}{!}{
\begin{tabular}{lccccccccc}
\toprule
  & \multicolumn{2}{c}{\%ALMs} & \multicolumn{2}{c}{ \%M20Ks} & \multicolumn{2}{c}{Avg Power (W)} &  \multicolumn{2}{c}{Latency (ms)} \\
\textbf{Core Count} & \textbf{Core} & \textbf{Total} & \textbf{Core} & \textbf{Total} & \textbf{Core Active} & \textbf{Core Idle} & \textbf{Core} & \textbf{DDR4}\\
\midrule
           1  & 2.9 &  9  & 0.01 & 2.87 & 13.67 & 13.68 & 46.36 & 0.09 \\
\midrule
          8  & 23.21 & 26.9 & 0.08 & 3.06 &  39.78  & 39.94 & 46.36 & 0.18 \\
          16  & 46.43 & 50.1 & 0.15 & 5.13 & 75.25 & 73.97 & 46.36 & 0.72 \\
          26  & 75.45 & 100 & 0.25 & 22.64 & 166.30 & 149.66 & 46.36 & 5.76 \\
\bottomrule \\
\end{tabular}
}
\label{tab:fpga-metrics1}
\end{table*}

\paragraph{The single core implementation} exhibits extremely low dynamic power that is hardly distinguishable from power measured while the core is inactive. Each core requires access to a DDR4 interface and MMIO bridges for host control. In this implementation, the majority of resources are dedicated to the provided shell logic and only 0.4\% of programmable logic resources are dedicated to logic for core interconnect and arbitration to DDR4 interfaces/MMIO. As described above, the core latency is primarily due to the larger execution time of the ternary matrix  multiply functional unit.

By instead using the full 512-bit DDR4 interface and parallelizing the TMATMUL functional unit, which dominates 99\% of core processing time, a further speed-up of approximately 64$\times$ is projected, while maintaining the same clock rate without additional optimizations or pipelining, as shown in \Cref{tab:fpga-metrics2}. Given the 370M parameter model where $L=24$, $d=512$, the total projected runtime is 16.08ms, and a throughput of approximately 62 tokens per second. The 1.3B parameter model, where $L=24$ and $d=2048$, has a projected runtime of 42ms, and a throughput of 23.8 tokens per second.
Despite not being optimized for 2-b precision weights, Intel Loihi~2 results outperform the FPGA implementation in terms of both throughput and power likely due to a mix of efficient packet routing, deep optimizations for recurrent state updates, unused digital signal processing (DSP) units on the FPGA, and process node deviations. 
% This reaches human reading speed at an efficiency that is on par with the power consumption of the human brain. 
This is for the case of a single core with a single batch of data, and can be scaled up significantly through batch processing by pipelining the single core with a negligible increase in average power, or alternatively, by increasing the core count with an increase in power (\Cref{tab:fpga-metrics1}).

\paragraph{Estimates of multi-core implementation latencies} are generated by scaling the overheads of the single core implementation and factoring in the growth of logic to accommodate contention on the DDR4 channels. Each core connects to one of four DDR4 channels, and each additional core connected to a channel will double the required arbitration and buffering logic for that channel. As both the host and core share DDR4 channels, this overhead will scale proportional to the number of cores attached to the channel. To mitigate this, future work could bring additional caching optimizations to the core and functional units. Core latency is the compute time of the core from start to ready and DDR4 latency is the required time to transfer input vectors from the host to the PAC local DDR4. 

\paragraph{Estimates of multi-core implementation power} are calculated by scaling the measured power of a single-core implementation. Idle power is estimated by scaling the total estimated resource overhead of all additional logic added to a constant estimate of idle power consumed by the platform shell. The single-core active power is scaled by the additional arbitration, interconnect and core overhead. We assume a constant clock rate for all implementations.

\begin{table*}[t]
\centering
\caption{FPGA Performance Metrics for Different Embedding Dimensions ($d$)}
\label{tab:fpga-metrics2}
\resizebox{\textwidth}{!}{
\begin{tabular}{lccccccc}
\toprule
 $d$ & \multicolumn{1}{c}{Runtime} & \multicolumn{1}{c}{Projected Runtime} & \multicolumn{2}{c}{Power (W)} & \multicolumn{2}{c}{ALM Utilization (\%)} & \multicolumn{1}{c}{Clock (MHz)} \\
 &  (ms) & w/Bursting (ms) & \textbf{Idle} & \textbf{Active} & \textbf{Core} & \textbf{Total} &  \\
\midrule
  512  & 43 & 0.67 & 13.36 & 13.39 & 2.8 & 9 & 60 \\
% \midrule
  1024 & 112 & 1.75 &  13.64 & 13.65 & 5.7 & 11 & 54 \\
  2048 & 456 & 7.13 & 13.92 & 13.93 & 11 & 16 & 52 \\
\bottomrule
\end{tabular}
}
\end{table*}

We note that the FPGA implementation is done in RTL from top to bottom, and there are many optimizations that could be added. For example, we are not using any vendor-provided IPs, and we are not bursting DDR transactions, both of which would significantly accelerate operation. This approach is to achieve the most generic and cross-platform evaluation possible. % We are currently working on a more optimized accelerator.

\section{Proof of Linear Scaling in the Proposed MatMul-free Linear Gated Recurrent Unit}

The MatMul-free Linear Gated Recurrent Unit (\name{}) presents a more efficient structure compared to the conventional Key-Value (KV) Cache mechanism used in Transformer models. To demonstrate its advantages, we examine both models in terms of computational complexity and memory requirements with respect to the sequence length \( T \).

In Transformers, the KV Cache mechanism stores all keys and values from previous time steps for autoregressive generation. At each time step \( t \), the model requires three steps: computing the query, key, and value vectors for the input \( \boldsymbol{x}_t \), followed by calculating the attention weights by comparing the query \( \boldsymbol{q}_t \) with all previous keys, and finally constructing the context vector by a weighted sum of the previous values. Formally, the query, key, and value are calculated as
\[
\boldsymbol{q}_t = \boldsymbol{x}_t \mathbf{W}_q, \quad \boldsymbol{k}_t = \boldsymbol{x}_t \mathbf{W}_k, \quad \boldsymbol{v}_t = \boldsymbol{x}_t \mathbf{W}_v,
\]
where \( \mathbf{W}_q, \mathbf{W}_k, \mathbf{W}_v \in \mathbb{R}^{d_{\text{model}} \times d_{\text{attn}}} \) are weight matrices. The attention weights, calculated as
\[
\alpha_{ti} = \frac{\exp\left(\frac{\boldsymbol{q}_t^\top \boldsymbol{k}_i}{\sqrt{d_{\text{attn}}}}\right)}{\sum_{j=1}^t \exp\left(\frac{\boldsymbol{q}_t^\top \boldsymbol{k}_j}{\sqrt{d_{\text{attn}}}}\right)}, \quad \text{for } i = 1, \dots, t,
\]
lead to the context vector
\[
\boldsymbol{z}_t = \sum_{i=1}^t \alpha_{ti} \boldsymbol{v}_i.
\]
These operations grow linearly in complexity with \( t \), producing \( O(t \cdot d_{\text{attn}}) \) computational steps per time step.

The computational complexity of the KV Cache over a sequence of length \( T \) therefore follows a quadratic scaling pattern:
\[
\sum_{t=1}^T O(t \cdot d_{\text{attn}}) = O(d_{\text{attn}} \cdot T^2).
\]
The memory requirement, dictated by the storage of all previous keys and values, also scales linearly with \( T \), resulting in \( O(T \cdot d_{\text{attn}}) \).

In contrast, the proposed \name{} architecture achieves simplicity and efficiency by eliminating matrix multiplications and relying on element-wise ternary-weight operations, making it a MatMul-free structure. This architecture uses a recurrent update rule for the hidden state \( \boldsymbol{h}_t \), leveraging the previous hidden state \( \boldsymbol{h}_{t-1} \) and the current input \( \boldsymbol{x}_t \) without retaining past sequence information. The key equations include the forget gate
\[
\boldsymbol{f}_t = \sigma\left(\boldsymbol{x}_t \circledast \mathbf{W}_f + \boldsymbol{b}_f\right),
\]
where \( \sigma \) represents the sigmoid function and \( \circledast \) denotes element-wise multiplication with ternary weights \( \mathbf{W}_f \in \{-1, 0, 1\}^{d \times d} \) and bias \( \boldsymbol{b}_f \in \mathbb{R}^d \). The candidate hidden state is given by
\[
\boldsymbol{c}_t = \tau\left(\boldsymbol{x}_t \circledast \mathbf{W}_c + \boldsymbol{b}_c\right),
\]
where \( \tau \) is the SiLU activation. The hidden state update, simplified through element-wise operations, is defined as
\[
\boldsymbol{h}_t = \boldsymbol{f}_t \odot \boldsymbol{h}_{t-1} + (1 - \boldsymbol{f}_t) \odot \boldsymbol{c}_t.
\]
The output computations, finalized by applying a sigmoid activation to the input and combining it with the hidden state, result in
\[
\boldsymbol{g}_t = \boldsymbol{x}_t \circledast \mathbf{W}_g + \boldsymbol{b}_g, \quad \boldsymbol{o}_t = \left(\boldsymbol{g}_t \odot \sigma (\boldsymbol{h}_t)\right) \circledast \mathbf{W}_o + \boldsymbol{b}_o.
\]
Each step requires only constant-time element-wise operations, establishing \( O(d) \) computational complexity per step.

Over a sequence of length \( T \), \name{} achieves total computational complexity of
\[
\sum_{t=1}^T O(d) = O(d \cdot T).
\]
Memory requirements remain constant at \( O(d) \) since only the current hidden state is maintained. Thus, \name{}’s efficiency is clearly shown by its linear computational complexity and constant memory demand, contrasting the quadratic growth and linear memory of the KV Cache.

The recursion in \name{} allows the hidden state \( \boldsymbol{h}_t \) to encapsulate all historical information without explicit storage, as shown by
\[
\boldsymbol{h}_t = \Phi(\boldsymbol{h}_{t-1}, \boldsymbol{x}_t),
\]
where \( \Phi \) represents the recursive update rule using element-wise ternary operations. This recursive nature ensures \( \boldsymbol{h}_t \) implicitly reflects the sequence \( \{\boldsymbol{x}_1, \dots, \boldsymbol{x}_t\} \) while avoiding the need for an accumulating memory footprint.

\section{Loihi 2: Detailed Implementation and Experimental Results}
\label{app:loihi_detailed_section}
This section of the appendix provides a comprehensive overview of the methodologies used for implementing the MatMul-free Language Model on Intel's Loihi 2 neuromorphic research chip, along with a detailed presentation of the experimental results. We elaborate on the relevant aspects of the Loihi 2 hardware architecture, the specifics of fixed-point conversion and custom function implementation critical for on-chip execution, and the performance benchmarks achieved, including throughput and energy efficiency. The subsequent subsections delve into these areas in greater detail.

\subsection{Loihi 2 Hardware Architecture}
\label{app:l2-hw-arch}

\begin{figure}[h!]
\centering
\includegraphics[width=0.55\linewidth]{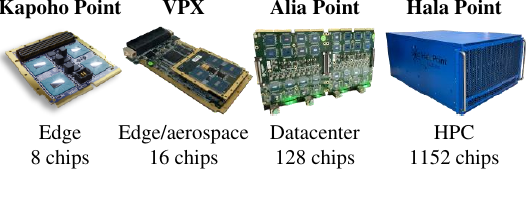}
\caption{Different Loihi 2 systems are available to cover a wide range of applications from the edge to HPC with up to 1 billion neurons.}
\label{fig:loihi_systems}
\end{figure}

Intel's second-generation neuromorphic research processor, Loihi 2, was purpose-built for sparse, event-based neural networks \citep{orchard2021efficient}. Each chip comprises 120 massively parallel compute units called \textit{neuro-cores} that can be scaled to systems containing up to 1,152 chips (Figure \ref{fig:loihi_systems}).
These neuro-cores operate asynchronously while maintaining global algorithmic time steps via barrier synchronization. Co-located memory enables efficient local state updates, supporting up to 8,192 stateful neurons per core. Users can program each neuron's temporal dynamics through assembly code, with flexible memory allocation achieved by trading neuron count for memory per core. External communication supports up to 160 million 32-bit integer messages per second \citep{shrestha_efficient_2024}. Fully-digital stacked systems like Hala Point can scale to 1 billion neurons and 128 billion synapses (Figure \ref{fig:loihi_systems}).
Networks must use fixed-point arithmetic: 8-bit synaptic weights\footnote{This is not a hard limit, as an $8n$-bit synapse can be implemented through
$n$ separate 8-bit synapses that are added together with different fixed-point exponents.} and up to 32-bit messages\footnote{Local states are not restricted in precision, and one may also transmit messages with more than 32 bits in an analogous way to what is described above for synaptic weights.}. Unlike GPUs, Loihi 2 prioritizes local neuronal computations—a defining characteristic of neuromorphic processors. This design enables: (1) fast, efficient recurrent state updates with minimal data movement through neuro-core-local memory; (2) efficient processing of unstructured sparse weight matrices via asynchronous event-driven architecture; and (3) exploitation of activation sparsity, as asynchronous communication transfers only non-zero messages.

\subsubsection{Execution Modes on Loihi 2}
\label{app:exmode}

\begin{figure}[h]
    \centering
    \includegraphics[width=0.5\linewidth]{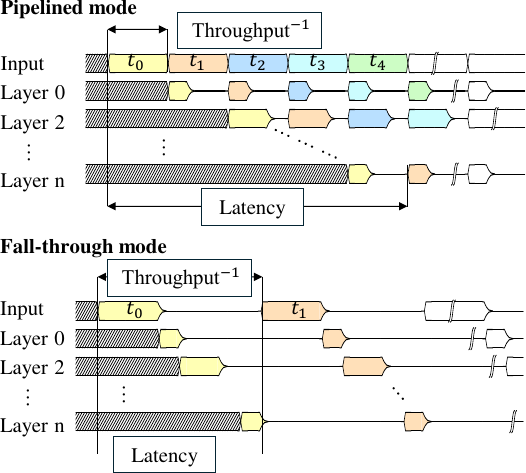}
    \caption{Different execution modes on Loihi 2 that either optimize throughput or latency. In the \textit{pipelined mode}, a new data point is inserted in each time step, to use all processing cores and maximize the throughput--at the expense of latency because equal time bins $t_0=t_1=\ldots$ are enforced. In the \textit{fall-through mode}, a new data points is only provided once the last data point has been fully processed with minimum latency. Only a single neuronal layer is active at any step as data travels through the network. The time per step is thus minimized as traffic is reduced and potentially more complex neuronal layers are not updated.}
    \label{fig:loihi_exec_modes}
\end{figure}

Loihi 2's asynchronous architecture enables dynamic throughput-latency optimization (Fig.~\ref{fig:loihi_exec_modes}). \textit{Pipelined mode} maximizes throughput by injecting new inputs every time step and forwarding them through neuronal layers. \textit{Fall-through mode} minimizes latency by introducing new inputs only after complete processing of previous inputs. LLM deployment naturally maps to these modes: prefill processing of long sequences utilizes pipelined mode for optimal throughput, while autoregressive generation employs fall-through mode since token generation at time $t$ must complete before processing token $t+1$.

\subsection{Fixed-Point Implementation Details}
\label{app:fxp-implementation}

\subsubsection{Fixed-Point Implementation of the Sigmoid Function}
\label{app:fxp-sigmoid}

We implement the logistic sigmoid function $\sigma(x) = 1/(1 + e^{-x})$ using fixed-point arithmetic with an interpolated look-up table (LUT). The floating-point input $x$ is scaled by $2^{x_{\text{exp}}}$ (where $x_{\text{exp}}=6$) and quantized to integer domain: $x_\text{fxp} = \lfloor x 2^{x_\text{exp}} \rceil$. The LUT stores precomputed values $\left(x_\text{int}^\text{lut,i} \mapsto y_\text{int}^\text{lut,i}\right)_{i \in \{0, \ldots N_\sigma\}}$ with $N_\sigma=8$ entries:
\begin{equation}
    y_\text{int}^\text{lut,i} = \lfloor \sigma\!\bigl(\frac{x^\text{lut,i}_\text{int}}{2^{x_{\text{exp}}}}\bigr)\cdot 2^{y_{\text{exp}}}\rfloor
\end{equation}
Only positive inputs are stored, as negative values are computed via $\sigma(-x) = 1 - \sigma(x)$, halving memory requirements. Piecewise linear interpolation between adjacent entries enhances output precision. Parameters $x_\text{exp}$ and $N_\sigma$ control the computation-accuracy tradeoff.

\subsubsection{Fixed point implementation of the inverse square root}
\label{app:fixed-pt-invsqrt}

For the inverse-square-root in the RMSNorm layer, we adapted a well-known “fast inverse square root” algorithm \texttt{FastInvSqrt} to operate entirely in fixed-point arithmetic.
Our method treats the input $\tilde{x}$ as an integer paired with a fixed exponent, and uses a LUT with 24 values to produce an initial guess for $\sqrt{\tilde{x}}$. This estimate is then refined using five iterations of the Newton-Raphson method, all in a fixed-point format.

\subsection{Double RMSNorm Derivation}
\label{app:double-rmsnorm}

Let $x \in \mathbb{R}^d$ and $y=\text{RMSNorm}(x)$ and $z=\text{RMSNorm}(y)$, in expanded form:
\begin{align}
    y &= \frac{x}{\sqrt{\epsilon + \sum_i^{d} x_i^2}} \odot g_1 \\
    z &= \frac{y}{\sqrt{\epsilon + \sum_i^{d} y_i^2}} \odot g_2
\end{align}
We wish to derive an equation for $z = \text{DoubleRMSNorm} (x) = \text{RMSNorm}(\text{RMSNorm}(x))$.

First, we express $\mu_{\mathbf{y}}$ in terms of $\mu_{\mathbf{x}}$:
\begin{align}
\mu_{\mathbf{y}} &= \frac{1}{D} \sum_{i=1}^D y_i^2 = \frac{1}{D} \sum_{i=1}^D \left( x_i \cdot \frac{g_1}{\sqrt{\mu_{\mathbf{x}} + \varepsilon}} \right)^2 \\
&= \left( \frac{g_1^2}{\mu_{\mathbf{x}} + \varepsilon} \right) \cdot \left( \frac{1}{D} \sum_{i=1}^D x_i^2 \right) = \frac{g_1^2 \mu_{\mathbf{x}}}{\mu_{\mathbf{x}} + \varepsilon}.
\end{align}
We then express $\mathbf{z}$ in terms of $\mathbf{x}$ by plugging in the equation for $\mathbf{y}$:
\begin{align}
\mathbf{z} &= \mathbf{y} \cdot \frac{g_2}{\sqrt{\mu_{\mathbf{y}} + \varepsilon}} = \left( \mathbf{x} \cdot \frac{g_1}{\sqrt{\mu_{\mathbf{x}} + \varepsilon}} \right) \cdot \frac{g_2}{\sqrt{\mu_{\mathbf{y}} + \varepsilon}} \\
&= \mathbf{x} \cdot \frac{g_1 g_2}{\sqrt{ (\mu_{\mathbf{x}} + \varepsilon)(\mu_{\mathbf{y}} + \varepsilon) }}.
\end{align}
We simplify the denominator:
\begin{align}
\sqrt{ (\mu_{\mathbf{x}} + \varepsilon)(\mu_{\mathbf{y}} + \varepsilon) } &= \sqrt{ (\mu_{\mathbf{x}} + \varepsilon) \cdot \frac{g_1^2 \mu_{\mathbf{x}}}{\mu_{\mathbf{x}} + \varepsilon} + \varepsilon } \\
&= \sqrt{ (\mu_{\mathbf{x}} + \varepsilon) \cdot \frac{ \mu_{\mathbf{x}} ( g_1^2 + \varepsilon ) + \varepsilon^2 }{ \mu_{\mathbf{x}} + \varepsilon } } \\
&= \sqrt{ \mu_{\mathbf{x}} ( g_1^2 + \varepsilon ) + \varepsilon^2 }.
\end{align}
We then derive the final expression for $\mathbf{z}$:
\begin{align}
\mathbf{z} &= \mathbf{x} \cdot \frac{ g_1 g_2 }{ \sqrt{ \mu_{\mathbf{x}} ( g_1^2 + \varepsilon ) + \varepsilon^2 } }.
\end{align}

This provides the combined RMSNorm operation with different scaling parameters $g_1$ and $g_2$.
By combining two RMSNorm operations with different scaling parameters, we arrive at a single normalization step:
\begin{equation}
\mathbf{z} = \mathbf{x} \cdot \frac{ g_{\text{combined}} }{ \sqrt{ \mu_{\mathbf{x}} + \varepsilon_{\text{combined}} } },
\end{equation}
where:
\begin{align}
g_{\text{combined}} = \frac{ g_1 g_2 }{ \sqrt{ g_1^2 + \varepsilon } }, \quad 
\varepsilon_{\text{combined}} = \frac{ \varepsilon^2 }{ g_1^2 + \varepsilon }.
\end{align}

Alternatively, since the denominator depends on $\mu_{\mathbf{x}}$, it may not be possible to express $\varepsilon_{\text{combined}}$ independently without further approximations.

\subsection{Detailed Hardware Results}
\label{app:detailed-hw-results}

Table \ref{tab:jetson-loihi-comp-ext} presents comprehensive performance comparisons between the MatMul-free LLM on Loihi 2 and transformer-based LLMs on NVIDIA Jetson Orin Nano, additionally including H100 GPU results for both MatMul-free LLM and Transformer++ baseline. 
The metrics for Loihi 2 are based on experiments on multi-chip systems. See Appendix \ref{app:l2-results-detailed} for more details.
The MatMul-free LLM on GPU demonstrates improved throughput and efficiency at longer sequences, attributed to linear token mixing scaling and enhanced GPU utilization. Conversely, the Transformer++ baseline exhibits initial throughput gains at shorter sequences through improved hardware utilization, followed by performance degradation at longer sequences due to self-attention's quadratic scaling.
The MatMul-free LLM on Loihi 2 shows constant scaling in both throughput and efficiency for different sequence lengths. This is due to full utilization of the chip at sequences as short as 500, and constant scaling of the MatMul-free token mixer.

\begin{table*}[h]
\caption{
Throughput and energy efficiency for two transformer-based language models running on the NVIDIA Jetson Orin Nano compared to our MatMul-free LM running on Intel's Loihi 2, across different sequence lengths for prefill and generation. 
The best-performing sequence length for each model and metric is \underline{underlined}.
% For more details regarding Loihi 2 metrics, see Appendix \ref{app:l2-results-detailed}. 
% \textbf{Gen}: autoregressive generation, \textbf{Prefill}: prefill mode. 
% $^*$ Llama representative model from \cite{alireo2024}.
}
\centering
\resizebox{\textwidth}{!}{
\begin{tabular}{clr|rrrrr|rrrrr}
\toprule
 & & & \multicolumn{5}{c}{Throughput ($\uparrow$ tokens/sec)} & \multicolumn{5}{c}{Efficiency ($\downarrow$ mJ/token)} \\
 & \multicolumn{2}{r}{Sequence length} & 500 & 1000 & 4000 & 8000 & 16000 & 500 & 1000 & 4000 & 8000 & 16000 \\
 
\midrule
\multirow{5}{*}{\rotatebox{90}{Generate}}
%%%%
& \textbf{MMF (370M)} & \textbf{Loihi 2$^\dagger$} & \textbf{59.4} & \textbf{59.4} & \textbf{59.4} & \textbf{59.4} & \textbf{59.4} & \textbf{70.8} & \textbf{70.8} & \textbf{70.8} & \textbf{70.8} & \textbf{70.8} \\
%%%%
& MMF (370M) & Loihi 2$^*$ & 71.3 & 71.3 & 71.3 & 71.3 & 71.3 & 59 & 59 & 59 & 59 & 59 \\
%%%%
% 24-chip results below:
 % & \textbf{MMF (370M)} & \textbf{Loihi 2} & \textbf{41.5} & \textbf{41.5} & \textbf{41.5} & \textbf{41.5} & \textbf{41.5} & \textbf{405} & \textbf{405} & \textbf{405} & \textbf{405} & \textbf{405} \\
%%%%
 & MMF (370M)          & H100     & 13.4       & 13.3 & \underline{13.5}    &  13.2   & \underline{13.5}  &  10.1k   & 10.1k & 10.0k    & 9.9k   &  \underline{9.8k} \\
 & TF++ (370M)          & H100    & 22.4       & \underline{22.9} & 21.7    &  21.3   & 20.9  &  \underline{5.5k}   & 5.6k  & 6.2k    & 6.8k   &  8.2k \\
 & Alireo (400M)     & Jetson  & 14.3    & 14.9 & 14.7 & \underline{15.2} & 12.8  & 723 & \underline{719}   & 853  & 812 & 1.2k \\
 & Qwen2 (500M)         & Jetson  & 13.4    & 14.0 & 14.1 & \underline{15.4} &  12.6     & 791 & \underline{785}   & 912  & 839 &  1.2k   \\

\midrule
\multirow{5}{*}{\rotatebox{90}{Prefill}}
%%%%
& \textbf{Ours (370M)} & \textbf{Loihi 2$^\dagger$} & \textbf{11637} & \textbf{11637} & \textbf{11637} & \textbf{11637} & \textbf{11637} & \textbf{3.4} & \textbf{3.4} & \textbf{3.4} & \textbf{3.4} & \textbf{3.4} \\
%%%%
& MMF (370M) & Loihi 2$^*$ & 13965 & 13965 & 13965 & 13965 & 13965 & 2.8 & 2.8 & 2.8 & 2.8 & 2.8 \\
%%%%
 % & \textbf{MMF (370M)} & \textbf{Loihi 2 (24-chip)} & \textbf{6632} & \textbf{6632} & \textbf{6632} & \textbf{6632} & \textbf{6632} & \textbf{3.7} & \textbf{3.7} & \textbf{3.7} & \textbf{3.7} & \textbf{3.7} \\
%%%%
 & MMF (370M)          & H100     & 11.4k    & 13.1k & 30.6k    &  51.6k   & \underline{84.6k}  & 6.1     & 5.3  & 2.5    & 1.4    & \underline{0.9}  \\
 & TF++ (370M)          & H100    & 21.6k    & 32.7k & 44.3k     & 55.4k   & \underline{60.5k}    & 11.3     & 7.3  & 5.4    & 4.3    & \underline{3.8}  \\
 & Alireo (400M)     & Jetson  & 849  & 1620  & \underline{3153} & 2258 & 1440   & 11.7  & 7.8  & \underline{6.8}  & 7.6  & 11.5  \\
 & Qwen2 (500M)         & Jetson  & 627  & 909   & 2639 & \underline{3861} & 3617   & 17.9  & 13.9 & 6.7  & \underline{4.4}  & 5.3 \\
\bottomrule
\multicolumn{13}{p{18.5cm}}{
\tiny$^*$ The MatMul-free LM on Loihi 2 was characterized on an Oheo Gulch single-chip Loihi 2 system (N3C1 silicon) running NxKernel v0.2.0 and NxCore v2.5.8 (only accessible to Intel Neuromorphic Research Community members). The 1-chip case neglects inter-chip communication.
\par
$^\dagger$ Inter-chip communication causes a derived $\approx$20\% slowdown over the single chip case (Appendix \ref{app:l2-results-detailed}).
\par
$^\ddagger$ Transformer LMs characterized on NVIDIA Jetson Orin Nano 8GB using MAXN power mode running Jetpack 6.2, TensorRT 10.3.0, CUDA 12.4. Energy values include CPU\_GPU\_CV, SOC, and IO components as reported by jtop 4.3.0.
\par
Performance results as of Jan 2025 and may not reflect all publicly available security updates. Results may vary.
}
\end{tabular}
}
\label{tab:jetson-loihi-comp-ext}
\end{table*}

\subsubsection{Detailed Loihi 2 Results}
\label{app:l2-results-detailed}

\paragraph{Single chip experiments}
As detailed in Section \ref{sec:loihi_results}, Loihi 2 energy and throughput metrics were derived from experiments deployed on a single MatMul-free LM block on an Oheo Gulch single-chip Loihi 2 system. We measured average time per step (TPS, $T_\text{TPS}$), representing single execution timestep duration. Total model latency (time-to-first-token, $T_\text{ttft}$) was calculated as $T_\text{ttft} = N_\text{blocks} \times N_\text{steps/block} \times T_\text{TPS}$, where $N_\text{blocks}=24$ for the 370M MatMul-free language model.

\begin{figure}[h]
    \centering
    \includegraphics[width=0.9\linewidth]{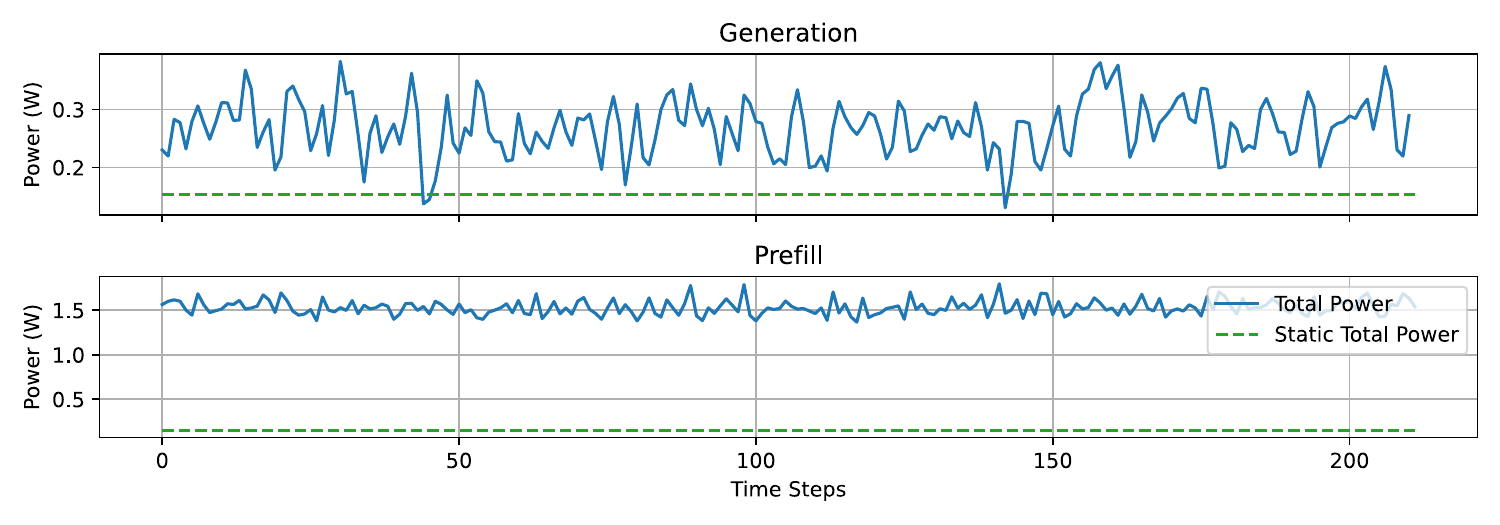}
    \caption{Power of one MatMul-free block on a single-chip Loihi 2 system.}
    \label{fig:l2-power-single-chip}
\end{figure}

Prefill operations employ pipelined mode with constant TPS due to enforced equal time bins (Appendix \ref{app:exmode}). Prefill throughput follows $f_\text{prefill} = T_\text{TPS}^{-1}$, as tokens are processed at interval $T_\text{TPS}$. Single-chip power measurements comprise static and dynamic components: $P^\text{1-chip}=\tilde P^\text{1-chip} + \bar P^\text{1-chip}$, where $\bar P$ and $\tilde P$ denote static and dynamic power respectively. Total prefill power estimates yield $\hat P_\text{prefill} = 24 \times P^\text{1-chip}$, with energy per token calculated as $\hat E_\text{prefill} = \hat P_\text{prefill} * T_\text{TPS}$. Fig.~\ref{fig:l2-power-single-chip} illustrates single-chip static and dynamic power measurements.

Generation utilizes fallthrough mode where TPS varies temporally, reflecting current operation latency (Appendix \ref{app:exmode}). Average TPS across $>$1000 timesteps determines $T_\text{TPS}$. Generation throughput equals $f_\text{generate} = T_\text{ttft}^{-1}$, typically lower than $f_\text{prefill}$. Power measurements follow prefill methodology, but full-model power estimation uses $\hat P_\text{generate} = \tilde P^\text{1-chip} + 24 \times \bar P^\text{1-chip}$, as only one chip processes information at any timestep while others remain idle, drawing only static power. Energy per token estimates follow $\hat E_\text{generate} = \hat P_\text{generate} * T_\text{ttft}$.

\paragraph{Multi-chip experiments}
\label{app:multi_chip_loihi}
Single-chip estimates inherently exclude inter-chip communication overhead. To address this limitation, we deployed all 24 MatMul-free LM blocks on the Alia Point system (Figure \ref{fig:loihi_systems}), utilizing 32 of 128 available chips. Each block maps to one chip, occupying 24 of 32 chips. Throughput calculations follow single-chip methodology, while power and energy metrics are directly measured. 
We measured the time-per-step (TPS) with increasing chip count on the Alia Point Loihi 2 system. The TPS remains stable for systems using up to four chips, then increases by $\approx 20\%$ and remains stable up to the full 24-chip system. Energy per token remains stable across all chip counts.
Accordingly, inter-chip communication is derived to cause an approximately $20\%$ slowdown over the single chip performance results (Table~\ref{tab:jetson-loihi-comp}).

This scaling behavior indicates favorable performance for larger MatMul-free models on Loihi 2 systems.
% Table \ref{tab:loihi-table-1-24-chip} compares single-chip estimates with full 24-chip implementation results.

% \begin{figure}[h]
%     \centering
%     \includegraphics[width=0.8\linewidth]{figures/l2_chip_scaling_prefill.pdf}
%     \caption{Scaling of time per step (inversely proportional to throughput, see text), power per chip and energy per token, as more chips are utilized in a 32-chip Alia Point Loihi 2 system. Each block of the MatMul-free LLM is implemented on a single Loihi 2 chip.}
%     \label{fig:l2-chip-scaling}
% \end{figure}

% Throughput decreased by $\approx 2.1 \times$ for prefill and $\approx 1.7 \times$ for generation, partially attributable to inter-chip communication. 
% Figure \ref{fig:l2-chip-scaling} illustrates TPS scaling with increasing chip count. Performance degradation stabilizes at $\geq 5$ chips, consistent with single activation vector transfers between chips per timestep (or every $N_\text{steps/block}$ steps in generation mode). This scaling behavior indicates favorable performance for larger MatMul-free models on Loihi 2 systems.
% Throughput decreased by $\approx 20\%$ while energy per token remains stable. 

\paragraph{Embedding and un-embedding layers} 
Current experiments exclude embedding and un-embedding layers mapping between $\mathbb{R}^{1024}$ embedding space and vocabulary size $V=32,000$. 
% Implementation plans include LUT-based embedding and ternary weight matrix ($1024 \times V$) un-embedding. 
Our experiments mapped un-embedding to 7 chips, totaling 31 chips for the complete 370M model. %, fitting within 32-chip constraints. 
The unembedding layer is partitioned and heavily parallelized such that throughput remains stable.
% Single-layer architecture minimizes latency impact; throughput should remain stable. 
% Additional power draw from these layers is expected to be offset by ongoing optimizations.

\subsubsection{Detailed NVIDIA Jetson Results}
\label{app:jetson-results-detailed}

We evaluated state-of-the-art transformer language models on NVIDIA Jetson Orin Nano 8GB (MAXN power mode, Jetpack 6.2, TensorRT 10.3.0, CUDA 12.4). Power measurements included GPU, SOC, and I/O subsystems via integrated profiling tools.
Two inference modes were evaluated: \textbf{prefill mode} (parallel input prompt processing) and \textbf{generate mode} (sequential autoregressive token generation). Generate mode's inherent token dependencies preclude sequence-level parallelization, yielding lower throughput than prefill mode.
Performance metrics—throughput (tokens/second), average power (Watts), and energy per token (Joules)—were derived from 30-iteration averaged measurements using \texttt{jtop} utility.

% \begin{figure}
%     \centering
%     \includegraphics[width=\linewidth]{figures/jetson_plot.pdf}
%     \caption{Hardware results for transformer-based LLMs running on the NVIDIA Jetson Orin Nano. \textit{Left}: prefill mode where text is ingested by the LLM. \textit{Right}: generate mode where text is generated in an auto-regressive loop. \textit{Top}: throughput in tokens per second. \textit{Middle}: average power in Watts. \textit{Bottom}: energy per token. All results are averaged over time for 30 inference runs. Results for the MatMul-free LLM running on Loihi 2 are based on estimates, as explained in Appendix \ref{app:l2-results-detailed}.}
%     \label{fig:jetson-results-figure}
% \end{figure}

Prefill mode achieved thousands of tokens/second through parallel token processing and effective pipelining. Generate mode demonstrated lower throughput due to sequential processing constraints. Power consumption varied between modes: prefill sustained higher continuous power draw, while generate exhibited variable profiles alternating between active processing and idle states.

Energy analysis revealed prefill mode's high throughput coincided with increased energy per token. Generate mode's extended latency elevated total energy per token despite lower instantaneous power draw.

These baseline measurements enable direct comparison with Loihi 2 neuromorphic implementations. Transformer models on Jetson Orin Nano achieve substantial throughput but at higher energy costs compared to MatMul-free LM estimates on Loihi 2, see Table \ref{tab:jetson-loihi-comp-ext}.

\begin{figure}
    \centering
    \includegraphics[width=0.9\linewidth]{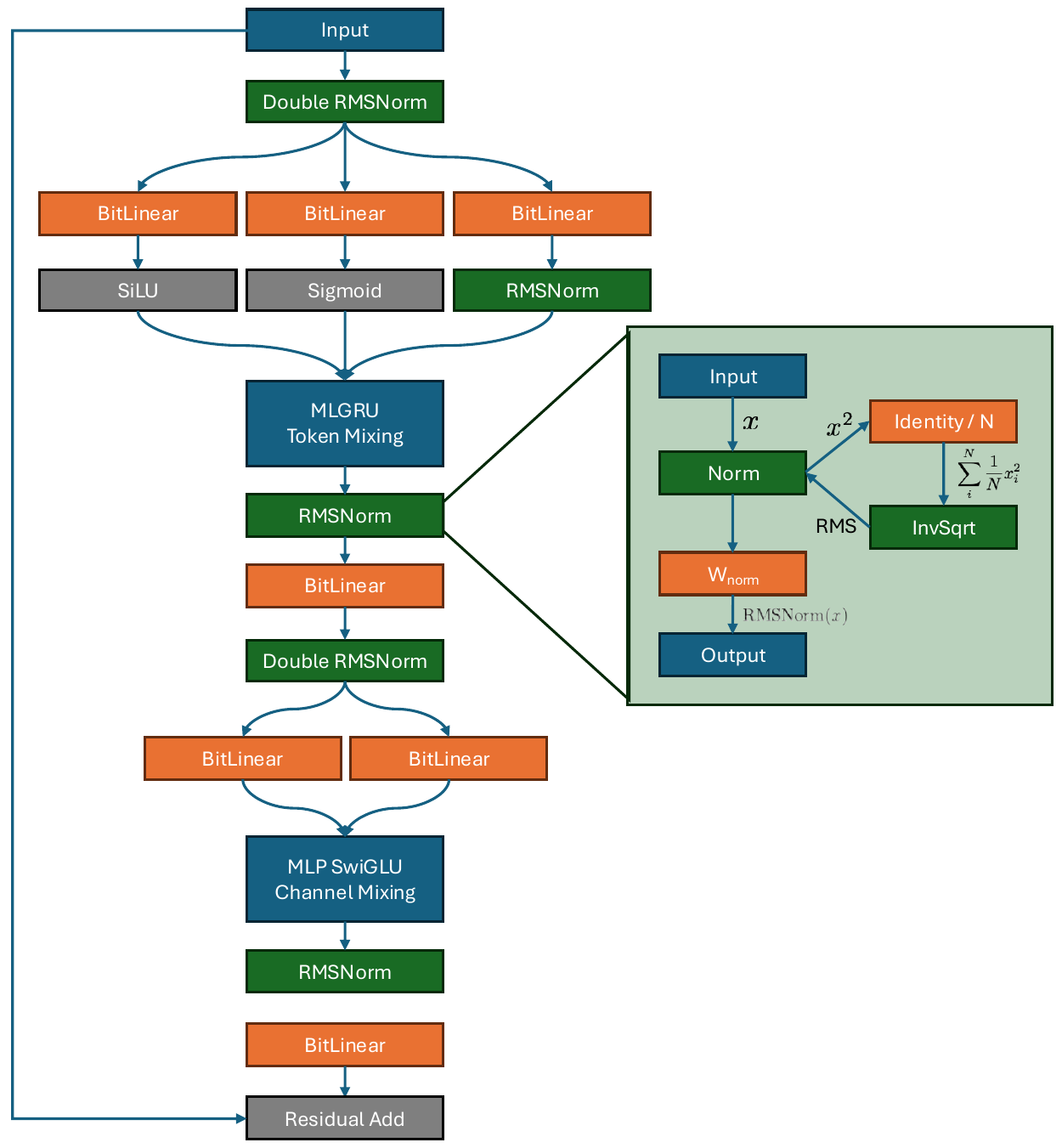}
    \caption{\textbf{Left}: Computational graph of a single MatMul-free LM layer, simplified from the actual computational graph that is mapped on the Loihi 2 chip. The RMSNorm is visualized as a single node. \textbf{Right}: Computational graph of the RMSNorm layer implemented on the Loihi 2 chip. See Sec.~\ref{ss:loihi-methods} for corresponding explanation.}
    % Neurons, synapses
    % Diagonal weights are not hidden for brevity. RMSNorm can have norm weights fused into neuron program, or separate. 
    % missing synapses are identity
    \label{fig:loihi-mmf-graph}
\end{figure}
\end{document}